\documentclass{article}

\usepackage[preprint,nonatbib]{neurips_2021}
\usepackage[
  style=authortitle,
  citestyle=authoryear,
  maxnames=2,
  maxbibnames=99,
  hyperref=true,
  backend=bibtex]{biblatex}

\DeclareNameAlias{sortname}{last-first}

\renewbibmacro{in:}{}

\DeclareCiteCommand{\cite}
  {\usebibmacro{prenote}}
  {\usebibmacro{citeindex}%
  \printtext[bibhyperref]{\usebibmacro{cite}}}
  {\multicitedelim}
  {\usebibmacro{postnote}}

\DeclareCiteCommand*{\cite}
  {\usebibmacro{prenote}}
  {\usebibmacro{citeindex}%
  \printtext[bibhyperref]{\usebibmacro{citeyear}}}
  {\multicitedelim}
  {\usebibmacro{postnote}}

\DeclareCiteCommand{\parencite}[\mkbibbrackets]
  {\usebibmacro{prenote}}
  {\usebibmacro{citeindex}%
    \printtext[bibhyperref]{\usebibmacro{cite}}}
  {\multicitedelim}
  {\usebibmacro{postnote}}

\DeclareCiteCommand*{\parencite}[\mkbibbrackets]
  {\usebibmacro{prenote}}
  {\usebibmacro{citeindex}%
    \printtext[bibhyperref]{\usebibmacro{citeyear}}}
  {\multicitedelim}
  {\usebibmacro{postnote}}

\DeclareCiteCommand{\footcite}[\mkbibfootnote]
  {\usebibmacro{prenote}}
  {\usebibmacro{citeindex}%
  \printtext[bibhyperref]{ \usebibmacro{cite}}}
  {\multicitedelim}
  {\usebibmacro{postnote}}

\DeclareCiteCommand{\footcitetext}[\mkbibfootnotetext]
  {\usebibmacro{prenote}}
  {\usebibmacro{citeindex}%
  \printtext[bibhyperref]{\usebibmacro{cite}}}
  {\multicitedelim}
  {\usebibmacro{postnote}}

\DeclareCiteCommand{\textcite}
  {\boolfalse{cbx:parens}}
  {\usebibmacro{citeindex}%
  \printtext[bibhyperref]{\usebibmacro{textcite}}}
  {\ifbool{cbx:parens}
     {\bibcloseparen\global\boolfalse{cbx:parens}}
     {}%
  \multicitedelim}
  {\usebibmacro{textcite:postnote}}

\usepackage[utf8]{inputenc} %
\usepackage[T1]{fontenc}    %
\usepackage{hyperref}       %
\usepackage{url}            %
\usepackage[ruled]{algorithm2e} 
\usepackage{subcaption}
\usepackage{graphicx}
\usepackage{booktabs}       %
\usepackage{amsmath, amsfonts}       %
\usepackage{nicefrac}       %
\usepackage{microtype}      %
\usepackage{xcolor}         %
\usepackage{amsthm}
\usepackage{tikz}
\usepackage{layouts}
\usepackage{array}
\usepackage{standalone}

\newtheorem{definition}{Definition}[section]

\usetikzlibrary{bayesnet}

\usepackage{ifthen}

\DeclareMathOperator*{\Exp}{\mathbb{E}}

\DeclareMathOperator{\Cov}{\textrm{Cov}}
\DeclareMathOperator{\KL}{\textrm{KL}}
\DeclareMathOperator{\midd}{||}

\newcommand{\dd}[2][]{\ensuremath{\frac{d #1}{d #2}}}

\newcommand{\pp}[2][]{\ensuremath{\frac{\partial #1}{\partial #2}}}

\newcommand{\too}{\longrightarrow}

\newcommand{\cZ}{\mathcal{Z}}

\newcommand{\e}{\epsilon}
\newcommand{\g}{\gamma}

\newcommand{\m}{\mu}

\renewcommand{\P}{\Phi}

\newcommand{\z}{\zeta}

\newcommand{\q}{q}

\renewcommand{\P}{\Pi}

\renewcommand{\g}{\gamma}

\renewcommand{\z}{z}

\renewcommand{\KL}{\mathrm{KL}}
\newcommand{\grad}{\nabla}

\newcommand{\f}{\ensuremath{\phi}}

\newcommand{\bigmm}{\ensuremath{\;\big|\big|\;}}

\newcommand{\Df}[2]{D_f\left( #1 \middle\Vert #2 \right)}
\renewcommand{\KL}[2]{\text{KL}\left( #1 \middle\Vert #2 \right)}
\newcommand{\pbv}{\phantom{\big\vert}}

\usepackage{color}
\usepackage{xcolor}
\definecolor{blue}{rgb}{0,0.2,0.5}
\definecolor{green}{rgb}{0.1,0.35,0.0}
\definecolor{red}{rgb}{0.5,0.0,0.0}
\definecolor{purple}{rgb}{0.4,0,0.6}
\definecolor{cyan}{rgb}{0.0,0.4,0.3}
\definecolor{orange}{rgb}{0.6,0.4,0.0}
\definecolor{gray}{rgb}{0.3,0.3,0.3}

\hypersetup{%
  colorlinks,
  linkcolor=red,
  citecolor=blue, %
  urlcolor=blue   %
}

\title{Nested Variational Inference}

\author{%
    Heiko Zimmermann$^{\dagger}$ \\
    \texttt{zimmermann.h@northeastern.edu}
    \And
    Hao Wu$^{\dagger}$ \\
    \texttt{wu.hao10@northeastern.edu}
    \AND
    Babak Esmaeili$^{\dagger}$ \\
    \texttt{esmaeili.b@northeastern.edu}
    \And
    Jan-Willem van de Meent$^{\dagger}$ \\
    \texttt{j.vandemeent@northeastern.edu}
    \And
    \normalfont
    $^{\dagger}$Khoury College of Computer Sciences,
    Northeastern University
}

\begin{document}
\maketitle

\begin{abstract}
    We develop nested variational inference (NVI), a family of methods that learn proposals for nested importance samplers by minimizing an forward or reverse KL divergence at each level of nesting. 
    NVI is applicable to many commonly-used importance sampling strategies and provides a mechanism for learning intermediate densities, which can serve as heuristics to guide the sampler. 
    Our experiments apply NVI to (a) sample from a multimodal distribution using a learned annealing path (b) learn heuristics that approximate the likelihood of future observations in a hidden Markov model and (c) to perform amortized inference in hierarchical deep generative models. We observe that optimizing nested objectives leads to improved sample quality in terms of log average weight and effective sample size. 
\end{abstract}

\begin{refsection}
\section{Introduction}
\label{sec:introduction}
Deep generative models provide a mechanism for incorporating priors into methods for unsupervised representation learning. This is particularly useful in settings where the prior defines a meaningful inductive bias that reflects the structure of the underlying domain. Training models with structured priors, however, poses some challenges.
A standard strategy for training deep generative models is to maximize a reparameterized variational lower bound with respect to a generative model and an inference model that approximates its posterior \parencite{kingma2013auto, rezende2014stochastic}. This approach works well in variational autoencoders with isotropic Gaussian priors, but often fails in models with high-dimensional and correlated latent variables. 

In recent years, a range of strategies for improving upon standard reparameterized variational inference have been put forward. 
These include wake-sleep style variational methods that minimize the forward KL-divergence \parencite{bornschein2015reweighted,le2019revisiting}, as well as sampling schemes that incorporate annealing \parencite{huang2018improving}, 
Sequential Monte Carlo \parencite{le2018auto-encoding,naesseth2017variational,maddison2017filtering}, 
Gibbs sampling \parencite{wu2019amortized, wang2018meta}, and MCMC updates \parencite{salimans2015markov, hoffman2017learning, li2017approximate}.
While these methods offer flexible inference, typically resulting in better approximations to the posterior compared to traditional variational inference methods, they are also model-specific, requiring specialized sampling schemes and gradient estimators, and can not be easily composed with other techniques.

In this paper, we propose nested variational inference, a framework for combining nested importance sampling and variational inference. Nested importance sampling formalizes the construction of proposals by way of calls to other importance samplers \parencite{naesseth2015nested, naesseth_elements_2019}. Many existing importance samplers are instances of nested samplers, including methods based on annealing \parencite{neal2001annealed} and sequential Monte Carlo \parencite{delmoral2006sequential}. 
NVI learns proposals by optimizing a divergence at each level of nesting. Combining nested variational objectives with importance resampling allows us to compute gradient estimates based on incremental weights, which depend only on variables that are sampled locally, rather than on all variables in the model. Doing so yields lower variance weights, whilst maintaining a high sample diversity relative to existing methods.

\section{Background}

\textbf{Stochastic Variational Inference.} Stochastic variational methods approximate a target density $\pi(z ; \theta) = \gamma(z ; \theta) / Z$ with parameters $\theta$ using a variational density $q(z ; \phi)$ with parameters $\phi$. Two common variational objectives are the forward and revserse Kullback-Leibler (KL) divergence, which are both instances of $f$-divergences,
\begin{align}
    \Df{\pi \pbv}{\pbv q}
    =
    \Exp_{q}
    \left[
      f\left(
      \frac{\pi(z ; \theta)}{q(z ; \phi)}
      \right)
    \right]
    =
    \begin{cases}
        \KL{\pi \pbv}{\pbv q} & f(\omega) =  \omega \log \omega, \\[4pt]
        \KL{q \pbv}{\pbv \pi} & f(\omega) =  - \log \omega.
    \end{cases}
\end{align}
Stochastic variational inference is commonly combined with maximum likelihood estimation for the parameters $\theta$. We are typically interested in the setting where $\pi(z;\theta)$ is the posterior density $p_\theta(z \,|\, x)$ of a model with latent variables $z$ and observations $x$. In this setting, $\gamma(z ; \theta) = p_\theta(x,z)$ is the joint density of the model, and $Z = p_\theta(x)$ is the marginal likelihood.

In the case of the reverse divergence $\KL{q\,}{\,\pi}$, also known as the \emph{exclusive} KL divergence, it is common practice to maximize a lower bound $\mathcal{L} = \Exp_{q}[\log (\gamma / q)] = \log Z - \KL{q \,}{\, \pi}$ with respect to $\theta$ and $\phi$. The gradient of $\mathcal{L}$ can be approximated using reparameterized samples \parencite{kingma2013auto,rezende2014stochastic}, likelihood-ratio estimators \parencite{wingate2013automated,ranganath2014black}, or a combination of the two \parencite{schulman2015gradient,ritchie2016deep}.

In the case of the forward divergence $\KL{\pi\,}{\,q}$, also known as the \emph{inclusive} KL divergence, stochastic variational methods typically optimize separate objectives for the inference and generative model. A common strategy is to approximate the gradients
\footnote{In order to optimize the parameters of the target density $\theta$, we can also maximize a lower bound on $\log Z$.}
\begin{align}
    \label{eq:grad-rws}
    \frac{\partial}{\partial \theta} \log Z 
    &= 
    \Exp_{\pi} \left[ 
      \frac{\partial}{\partial \theta} \log \gamma(z ; \theta) 
    \right]
    ,
    &
    -\frac{\partial}{\partial \phi} \: \KL{\pi \pbv}{\pbv q} 
    &=
    \Exp_{\pi} \left[ 
      \frac{\partial}{\partial \phi} \log q(z ; \phi) 
    \right]
    .
\end{align}
This requires samples from $\pi$, which itself requires approximate inference. A common strategy, which was popularized in the context of reweighted wake-sleep (RWS) methods \parencite{bornschein2015reweighted,le2019revisiting}, is to use $q$ as a proposal in an importance sampler.

\textbf{Self-Normalized Importance Samplers.} An expectation $\Exp_\pi[g(z)]$ with respect to $\pi$ can be rewritten with respect to a proposal density $q$ by introducing an unnormalized importance weight $w$,
\begin{align}
    \label{eq:self_normalized_is_estimator}
    \Exp_\pi \big[ g(z) \big] 
    &= 
    \frac{1}{Z} \Exp_q \big[w \: g(z) \big],
    &
    w
    =
    \frac{\gamma(z ; \theta)}
         {q(z ; \phi)}
    .
\end{align}
Self-normalized estimators use weighted samples $\{w^s, z^s\}_{s=1}^S$ to both approximate the expectation with respect to $q$, and to compute an estimate $\hat{Z}$ of the normalizer,
\begin{align}
    \Exp_\pi[g(z)]
    &\simeq
    \hat g = 
    \frac{1}{\hat{Z}}
    \frac{1}{S}
    \sum_{s=1}^S
    w^s \: g \big(z^s\big),
    &
    \hat{Z}
    &= 
    \frac{1}{S}
    \sum_{s=1}^S    
    w^s, 
    &
    z^s &\sim q(\cdot;\f).
\end{align}
This estimator is consistent, i.e. $\hat g \stackrel{a.s.}{\too} \Exp_\pi[g(z)]$ as the number of samples $S$ increases. 
However it is not unbiased, since it follows from Jensen's inequality that 
\begin{align}
    \frac{1}{Z} = \frac{1}{\Exp_q[\hat{Z}]} \le \Exp_q [\frac{1}{\hat{Z}}].
\end{align}
The degree of bias depends on the variance of the importance weight. When $q = \pi$, the importance weight $w=Z$ has zero variance, and the inequality 
is tight. 
In the context of stochastic variational inference, this means that gradient estimates might initially be strongly biased, since there will typically be a large discrepancy between $q$ and $\pi$. 
However, the variance will typically decrease as the quality of the approximation improves. This intuition can be made exact when optimizing a Pearson $\chi^2$-divergence, which corresponds to a variance minimizing objective, i.e. minimizing the variance of the corresponding importance weight \parencite{muller_neural_2019}. Here we focus on optimizing a KL divergences only, but still take motivation from this intuition.

\section{Nested Variational Inference}

A widely used strategy in importance sampling is to decompose a difficult sampling problem into a series of easier problems.
A common approach is to define a sequence of unnormalized densities $\{\gamma_k\}_{k=1}^K$ that interpolate between an initial density $\pi_1 = \gamma_1 / Z_1$, for which sampling is easy, and the final target density $\pi_K = \gamma_K / Z_K$. 
At each step, samples from the preceding density serve to construct proposals for the next density, which is typically combined with importance resampling or application of a Markov chain Monte Carlo (MCMC) operator to improve the average sample quality. 

\begin{figure}
    \centering
    \includegraphics[width=0.87\linewidth]{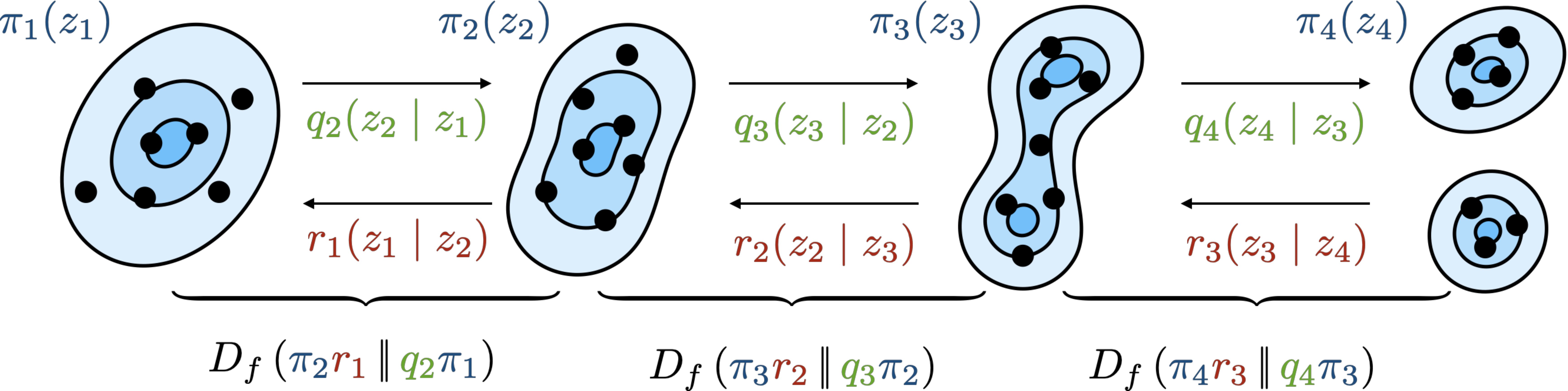}
    \caption{Nested variational inference minimizes an $f$-divergence at each step in a sequence of densities to learn forward proposals $q_k$, reverse kernels $r_{k-1}$, and intermediate densities $\pi_k$.}
    \label{fig:nvi-overview}
\end{figure}

NVI defines objectives for optimizing importance samplers that target a sequence of densities. 
To do so, at every step, it minimizes the discrepancy between a \emph{forward density} $\hat{\pi}_k = \hat{\gamma}_k / Z_{k-1}$, which acts as the proposal, and a \emph{reverse density} $\check{\pi}_k = \check{\gamma}_k / Z_k$, which defines an intermediate target. 
We define the forward density by combining the preceding target $\gamma_{k-1}$  with a forward kernel $q_k$, and the reverse density by combining the next target $\gamma_k$ with a reverse kernel $r_{k-1}$,
\begin{align}
    \check{\gamma}_k(z_k, z_{k-1})
    &=
    \gamma_k(z_k) \: r_{k-1}(z_{k-1} | z_k ),
    &
    \hat{\gamma}_k(z_k, z_{k-1})
    &=
    q_k(z_k | z_{k-1}) \: \gamma_{k-1}(z_{k-1}).
\end{align}

Our goal is to learn pairs of densities $\check{\pi}_k$ and $\hat{\pi}_k$ that are as similar as possible. To do so, we minimize a variational objective $\mathcal{D}$ that comprises an $f$-divergence for each step in the sequence, along with a divergence between the first intermediate target $\pi_1$ and an initial proposal $q_1$,
\begin{align}
    \mathcal{D}
    &=
    D_f
    \big(
    \pi_1%
    \bigmm
    \q_1%
    \big)
    +
    \sum_{k=2}^K
    D_f
    \big(
    \check{\pi}_k%
    \bigmm
    \hat{\pi}_k%
    \big)
    .
\end{align}
We can optimize this objective in two ways. The first is to optimize with respect to $q_1$ and the forward densities $\hat{\pi}_k$. This serves to learn proposals $q_k$, but can also be used to learn intermediate densities $\pi_{k}$ that are as similar as possible to the next density $\pi_{k+1}$. In settings where we wish to learn reverse kernels $r_{k}$, we can additionally minimize $\mathcal{D}$ with respect to the reverse densities $\check{\pi}_k$. Since each intermediate density $\pi_{k}$ occurs in both $\check{\pi}_k$ and in $\hat{\pi}_{k+1}$, this defines a trade-off between maximizing the similarity to  $\pi_{k+1}$ and the similarity to $\pi_{k-1}$.

\subsection{Nested Importance Sampling and Proper Weighting} 

Sampling from a sequence of densities can be performed using a nested importance sampling construction \parencite{naesseth2015nested, naesseth_elements_2019}, which uses weighted samples from $\pi_{k-1}$ as proposals for $\pi_{k}$. The technical requirements for such constructions can be summarized as follows:

\begin{definition}[Proper weighting]
    \label{def:proper_weighting}
    \textit{
    Let $\pi$ be a probability density. 
    For some constant $c > 0$,
    a random pair $(w, z) \sim \P$ is properly weighted (p.w.) for an unnormalized
    probability density $\gamma \equiv Z \pi$ if $w \geq 0$ and for all measurable functions $g$ it holds that
    }
    \begin{align*}
        \Exp_{z, w \sim \P}
        \left[ 
            w \: g(z) 
        \right] 
        = 
        c 
        \int dz\ \gamma(z) \: g(z) 
        = 
        cZ
        \Exp_{z \sim \pi}\left[g(z) \right]
        .
    \end{align*}
\end{definition} 

Given a pair $(w_{k-1}, z_{k-1})$ that is properly weighted for $\gamma_{k-1}$, we can use a sequential importance sampling construction to define a pair $(w_k, z_k)$ that is properly weighted for $\gamma_k$,
\begin{align}
    \label{eq:proper-weighting}
    z_k \sim q_k(\cdot \mid z_{k-1}),
    &&
    w_k 
    &= 
    v_k \, w_{k-1},
    &
    v_k
    =
    \frac{\check{\gamma}_k(z_k, z_{k-1})}
         {\hat{\gamma}_k(z_k, z_{k-1})}.
\end{align}
We refer to the ratio $v_k$ as the incremental weight. In this construction, $(w_k, z_{k-1:k})$ is properly weighted for $\check{\gamma}_k$ which implies that $(w_k, z_k)$ is also properly weighted for $\gamma_k$, since 
\begin{align}
    \int d z_{k-1} \: \check{\gamma}_k(z_k, z_{k-1}) = \int d z_{k-1} \: \gamma_k(z_k) r_{k-1}(z_{k-1} \,|\, z_{k}) = \gamma_k(z_k).
\end{align}
Sequential importance sampling can be combined with other operations that preserve proper weighting, including rejection sampling, application of an MCMC transition operator, and importance resampling. This defines a class of samplers that admits a many popular methods as special cases, including sequential Monte Carlo (SMC) \parencite{doucet2001sequential}, annealed importance sampling (AIS) \parencite{neal2001annealed}, and SMC samplers \parencite{chopin2002sequential}. 

These samplers vary in the sequences of densities they define. SMC for state-space models employs the filtering distribution on the first $k$ points in a time series as the intermediate target $\pi_k$. In this setting, where the dimensionality of the support increases at each step, we typically omit the reverse kernel $r_{k-1}$. In AIS and SMC samplers, where the support is fixed, a common strategy is to define an annealing path $\gamma_k(z_k) = \gamma_K(z_k)^{\beta_k} \gamma_1(z_k)^{1-\beta_k}$ that interpolates between the initial density $\gamma_1$ and the final target $\gamma_{K}$ by varying the coefficients $\beta_k$ from 0 to 1. 

\subsection{Computing Gradient Estimates} 
\label{sec:computing-gradients}

The NVI objective can be optimized with respect to three sets of densities. We will use $\theta_k$, $\hat{\phi}_k$, and $\check{\phi}_k$ to denote the parameters of the densities $\pi_k$, $q_k$, and $r_{k}$ respectively. 
For notational convenience, we use $\check{\rho}_k$ to refer to the parameters of the reverse density $\check{\pi}_k$, and $\hat{\rho}_k$ to refer to the parameters of the forward density $\hat{\pi}_k$,
\begin{align}
    \check{\rho}_k
    &=
    \{\theta_k, 
      \check{\phi}_{k-1}\}, 
    &
    \hat{\rho}_k 
    &= 
    \{\hat{\phi}_{k}, 
      \theta_{k-1}\}
    .
\end{align}

\textbf{Gradients of the Forward KL divergence.} When we employ the forward KL as the objective, the derivative with respect to $\hat{\rho}_k$ can be expressed as (see Appendix~\ref{sec:apx-grad-forward-kl}),
\begin{align}
    \label{eq:grad-fkl-fwd}
    -\frac{\partial}
          {\partial \hat{\rho}_k}
    \KL{\check{\pi}_k \!\pbv}{\pbv\! \hat{\pi}_k }
    &=
    \Exp_{\check{\pi}_k}
    \!
    \left[
        \frac{\partial}
             {\partial \hat{\rho}_k}
        \!
        \log \hat{\gamma}_k\big(z_k, z_{k-1} ; \hat{\rho}_k \big)
    \right]
    -
    \Exp_{\pi_{k-1}}
    \!
    \left[
        \frac{\partial}
             {\partial \hat{\rho}_k}
        \!
        \log \gamma_{k-1}\big(z_{k-1} ; \theta_{k-1} \big)
    \right]
    .
\end{align}
This case is the nested analogue of RWS-style variational inference. We can move the derivative into the expectation, since $\check{\pi}_k$ does not depend on $\hat{\rho}_k$. We then decompose $\log \hat{\pi}_k = \log \hat{\gamma}_k - \log Z_{k-1}$ and use the identity from Equation~\ref{eq:grad-rws} to express the gradient $\log Z_{k-1}$ as an expectation with respect to $\pi_{k-1}$. The resulting  expectations can be approximated using self-normalized estimators based on the outgoing weights $w_k$ and incoming weights $w_{k-1}$ respectively.

The gradient of the forward KL with respect to $\check{\rho}_k$ is more difficult to approximate, since the expectation is computed with respect to $\check{\pi}_k$, which depends on the parameters $\check{\rho}_k$. The gradient of this expectation has the form (see Appendix~\ref{sec:apx-grad-forward-kl})
\begin{align}
    &
    -\frac{\partial}
          {\partial \check{\rho}_k}
    \KL{\check{\pi}_k \pbv}{\pbv \hat{\pi}_k}
    =
    -
    \Exp_{\check{\pi}_k}
    \left[
      \log v_k \:
      \frac{\partial}
           {\partial \check{\rho}_k}
      \log \check{\pi}_k
           \big(z_k, z_{k-1} ; \check{\rho}_k \big)
    \right],
    \\
    \nonumber
    &\qquad
    =
    -
    \Exp_{\check{\pi}_k}
    \left[
      \log v_k \:
      \frac{\partial}
           {\partial \check{\rho}_k}
      \log \check{\gamma}_k
           \big(z_k, z_{k-1} ; \check{\rho}_k \big)
    \right] 
    +
    \Exp_{\check{\pi}_k}
    \left[
      \vphantom{\frac{\partial}{\partial}}
      \log v_k \:
    \right]
    \Exp_{\pi_k}
    \left[
      \frac{\partial}
           {\partial \check{\rho}_k}
      \log \gamma_k\big(z_k ; \theta_k \big)
    \right].
\end{align}
In principle, we can approximate this gradient using self-normalized estimators based on the outgoing weight $w_k$. We experimented extensively with this estimator, but unfortunately we found it to be unstable, particularly for the gradient with respect to the parameters of the reverse kernel $r_{k-1}$. For this reason, our experiments employ the reverse KL when learning reverse kernels.

Our hypothesis is that the instability in this estimator arises because the gradient \emph{decreases} the probability of high-weight samples and \emph{increases} the probability of low-weight samples, rather than the other way around.
This could lead to problems during early stages of training, when the estimator will underrepresent low-weight samples, for which the probability should increase.

\textbf{Gradients of the Reverse KL divergence.} When computing the gradient of the reverse KL with respect to $\hat{\rho}_k$, we obtain the nested analogue of methods that maximize a lower bound. Here we can either use reparameterized samples \parencite{kingma2013auto,rezende2014stochastic} or likelihood-ratio estimators \parencite{wingate2013automated,ranganath2014black}. We will follow \textcite{ritchie2016neurally} and define a unified  estimator in which proposals are generated using a construction 
\begin{align}
    w_k &= v_k w_{k-1},
    &
    z_k &= g(\tilde{z}_k, \hat{\phi}_k),
    &
    \tilde{z}_k &\sim \tilde{q}_k(\tilde{z}_k \mid z_{k-1}, \hat{\phi}_k),
    &
    w_{k-1}, z_{k-1} \sim \Pi_{k-1}.
\end{align}
This construction recovers reparameterized samplers in the special case when $\tilde{q}_k$ does not depend on  parameters, and recovers non-reparameterized samplers when $z_k = \tilde{z}_k$. This means it is applicable to models with continuous variables, discrete variables, or a combination of the two. The gradient of the reverse KL for proposals that are constructed in this manner becomes (see Appendix~\ref{sec:apx-grad-reverse-kl})
\begin{align}
    \nonumber
    -\frac{\partial}
          {\partial \hat{\rho}_k}
    \KL{\hat{\pi}_k \pbv}{\pbv \check{\pi}_k}
    &
    =
    \Exp_{\check{\pi}_k}
    \left[
        \frac{\partial}
             {\partial z_k}
        \log \hat{\gamma}_k\big(z_k, z_{k-1} ; \hat{\rho}_k \big)
        \frac{\partial z_k}
             {\partial \hat{\rho}_k}
    \right]
    \\
    &\qquad
    +
    \Exp_{\check{\pi}_k}
    \left[
        \log v_k
        \frac{\partial}
             {\partial \hat{\rho}_k}
        \log \hat{\gamma}_k\big(z_k, z_{k-1} ; \hat{\rho}_k \big)                    
    \right]
    \\
    \nonumber
    &\qquad
    - 
    \Exp_{\check{\pi}_k}
    \left[
        \vphantom{\frac{\partial}{\partial \hat{\rho}_k}}
        \log v_k
    \right]
    \Exp_{\pi_{k-1}}
    \left[
        \frac{\partial}
             {\partial \hat{\rho}_k}
        \log \gamma_{k-1}\big(z_{k-1} ; \theta_{k-1} \big)
    \right]
    .
\end{align}
In this gradient, the first term represents the pathwise derivative with respect to reparameterized samples. The second term defines a likelihood-ratio estimator in terms of the unnormalized density $\hat{\gamma}_k$, and the third term computes the contribution of the gradient of the log normalizer $\log Z_{k-1}$.

Computing the gradient of the reverse KL with respect to $\check{\rho}_k$ is once again straightforward, since we are computing an expectation with respect to $\hat{\pi}_k$, which does not depend on $\check{\rho}_k$. This means we can move the derivative into the expectation, which yields a gradient analogous to that in Equation~\ref{eq:grad-fkl-fwd},
\begin{align}
    -\frac{\partial}
          {\partial \check{\rho}_k}
    \KL{\hat{\pi}_k \pbv}{\pbv \check{\pi}_k }
    &=
    \Exp_{\hat{\pi}_k}
    \left[
      \frac{\partial}
           {\partial \check{\rho}_k}
      \log \check{\gamma}_{k}\big(z_k, z_{k-1} ; \check{\rho}_{k} \big)    
    \right]    
    -
    \Exp_{\pi_k}
    \left[
      \frac{\partial}
           {\partial \check{\rho}_k}
      \log \gamma_{k}\big(z_k ; \theta_{k} \big)    
    \right]    
    .
\end{align}

\textbf{Variance Reduction.} To reduce the variance of the gradient estimates we use the expected log-incremental weight as a baseline for score function terms and employ the sticking-the-landing trick \parencite{roeder2017sticking} when reparameterizing the forward kernel as described in Appendix~\ref{sec:apx-all-grad-estimate}.

\subsection{Relationship to Importance-Weighted and Self-Normalized Estimators} 

There exists a large body of work on methods that combine variational inference with MCMC and importance sampling. We refer to Appendix~\ref{sec:apx-related-work} for a comprehensive discussion of related and indirectly related approaches. To position NVI in the context of the most directly related work, we here focus on commonly used importance-weighted and self-normalized estimators.

NVI differs from existing methods in that it defines an objective pairs of variables $(z_k,z_{k-1})$ at each level of nesting, rather than a single objective for the entire sequence of variables $(z_1, \dots, z_K)$. One of the standard approaches for combining importance sampling and variational inference is to define an ``importance-weighted'' stochastic lower bound $\hat{\mathcal{L}}_K$ \parencite{burda2016importance},
\begin{align*}
    \hat{\mathcal{L}}_K &= \log \hat{Z}_K,
    &
    \hat{Z}_K = \frac{1}{S} \sum_{s=1}^S w^s_K.
\end{align*}
By Jensen's inequality, $\Exp[\hat{\mathcal{L}}_K] \le \log \Exp[\hat{Z}_K] = \log Z_K$, which implies that we can define a stochastic lower bound using any properly-weighted sampler for $\gamma_K$, including samplers based on SMC \parencite{le2018auto-encoding,naesseth2018variational,maddison2017filtering}. For purposes of learning the target density $\gamma_K$, this approach is equivalent to computing an RWS-style estimator of the gradient in Equation~\ref{eq:grad-rws},
\begin{align}
    \label{eq:nvi-sis}
    \frac{\partial}{\partial \theta_K}
    \hat{\mathcal{L}}_K
    =
    \frac{1}{\hat{Z}_K}
    \frac{1}{S}
    \sum_{s=1}^S
    w^s_K
    \frac{\partial}{\partial \theta_K}
    \log \gamma_K(z_K^s ; \theta_K)
    .
\end{align}
However, these two approaches are not equivalent for purposes of learning the proposals. We can maximize a stochastic lower bound to learn  $q_k$, but this requires doubly-reparameterized estimators \parencite{tucker2018doubly} in order to avoid problems with the signal-to-noise ratio in this estimator, which can paradoxically deteriorate with the number of samples \parencite{rainforth2018tighter}. The estimators in NVI do not suffer from this problem, since we do not compute the logarithm of an average weight.

NVI is also not equivalent to learning proposals with RWS-style estimators. If we use sequential importance sampling (SIS) to generate samples, a self-normalized gradient for the parameters of $q_k$ that is analogous to the one in Equation~\ref{eq:grad-rws} has the form
\begin{align}
    &
    \Exp_{\pi_K, r_{K-1}, \dots, r_1}
    \left[
    \frac{\partial}{\partial \hat{\phi}_k}        
    \log q_k(z_k \,|\, z_{k-1} \,;\, \hat{\phi}_k)
    \right]
    \simeq
    \frac{1}{\hat{Z}_K}
    \frac{1}{S}
    \sum_{s=1}^S
    w^s_K
    \frac{\partial}{\partial \hat{\phi}_k}        
    \log q_k(z^s_k \,|\, z^s_{k-1} \,;\, \hat{\phi}_k)
    .
\end{align}
Note that this expression depends on the final weight $w_K$. By contrast, a NVI objective based on the  forward KL yields a self-normalized estimator that is defined in terms of the intermediate weight $w_k$
\begin{align}
    &
    \Exp_{\pi_k, r_{k-1}}
    \left[
    \frac{\partial}{\partial \hat{\phi}_k}        
    \log q_k(z_k \,|\, z_{k-1} \,;\, \hat{\phi}_k)
    \right]
    \simeq
    \frac{1}{\hat{Z}_k}
    \frac{1}{S}
    \sum_{s=1}^S
    w^s_k
    \frac{\partial}{\partial \hat{\phi}_k}        
    \log q_k(z^s_k \,|\, z^s_{k-1} \,;\, \hat{\phi}_k)
    .
\end{align}
If instead of SIS we employ sequential importance resampling (i.e.~SMC), then the incoming weight $w_{k-1}$ is identical for all samples. This means that we can express this estimator in terms of the incremental weight $v_k$ rather than the intermediate weight $w_k$
\begin{align}
    &
    \Exp_{\pi_k, r_{k-1}}
    \left[
    \frac{\partial}{\partial \hat{\phi}_k}        
    \log q_k(z_k \,|\, z_{k-1} \,;\, \hat{\phi}_k)
    \right]
    \simeq
    \sum_{s=1}^S
    \frac{v^s_k}
         {\sum_{s'=1}^S v^{s'}_k}
    \frac{\partial}{\partial \hat{\phi}_k}        
    \log q_k(z^s_k \,|\, z^s_{k-1} \,;\, \hat{\phi}_k)
    .
\end{align}
We see that NVI allows us to compute gradient estimates that are localized to a specific level of the sampler. In practice, this can lead to lower-variance gradient estimates. 

Having localized gradient computations also offers potential memory advantages. Existing methods typically perform reverse-mode automatic differentiation on an objective that is computed from the final weights (e.g.~the stochastic lower bound). This means that memory requirements scale as $\mathcal{O}(SK)$ since the system needs to keep the entire computation graph in memory. In NVI, gradient estimates at level $k$ do not require differentiation of the incoming weights $w_{k-1}$, This means that it is possible to perform automatic differentiation on a locally-defined objective before proceeding to the next level of nesting, which means that memory requirements would scale as $\mathcal{O}(S)$. It should therefore in principle be possible to employ a large number of levels of nesting $K$ in NVI, although we do not evaluate the stability of NVI at large $K$ in our experiments.
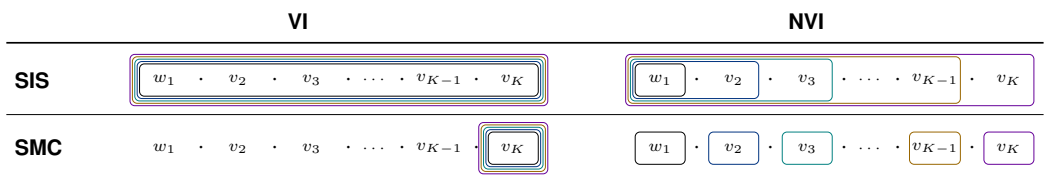
\begin{figure}[t]
    \centering
    \tikzset{
    fbox/.style={
        inner sep=1,%
        outer sep=0,%
        rounded corners=2pt,
    },
    wbox/.style={
        rounded corners=2pt,
        inner sep=0,%
        outer sep=0,%
        minimum width={width("\scriptsize$w_{k-1}$")},
        minimum height={height("\scriptsize$w_{k-1}$") + 3mm},
        font=\scriptsize
    },
    cdot/.style={
        inner sep=0,%
        outer sep=0,%
    },
    dots/.style={
        inner sep=0,%
        outer sep=0,%
        font=\tiny
    },
}
\tikzset{
    weights/.pic={
        \node[wbox](-v1) {$w_1$};
        \node[cdot, right=1.6mm of -v1.east, anchor=west](-cdot) {$\cdot$};
        \node[wbox, right=1.6mm of -cdot.east, anchor=west](-v2) {$v_2$};
        \node[cdot, right=1.6mm of -v2.east, anchor=west](-cdot) {$\cdot$};
        \node[wbox, right=1.6mm of -cdot.east, anchor=west](-v3) {$v_3$};
        \node[cdot, right=1.6mm of -v3.east, anchor=west](-cdot) {$\cdot$};
        \node[dots, right=1.6mm of -cdot.east, anchor=west](-dots) {$\cdots$};
        \node[cdot, right=1.6mm of -dots.east, anchor=west](-cdot) {$\cdot$};
        \node[wbox, right=1.6mm of -cdot.east, anchor=west](-vK-1) {$v_{K-1}$};
        \node[cdot, right=1.6mm of -vK-1.east, anchor=west](-dots) {$\cdot$};
        \node[wbox, right=1.6mm of -dots.east, anchor=west](-vK) {$v_{K}$};
    },
}
\begin{tikzpicture}
    \draw[line width=0.25mm] (-2.3, 0) -- (13, 0);
    \node[anchor=north] (vi-text) at (2.0, -0.1) {\small\textbf{\textsf{VI}}};
    \node[anchor=north] (nvi-text) at (9.5, -0.1) {\small\textbf{\textsf{NVI}}};
    \draw[line width=0.25mm] (-2.3, -0.65) -- (13, -0.65);
    \node[align=left, anchor=west](sis-text) at (-2.3, -1.2) {\small\textbf{\textsf{SIS}}};
    \pic[right=2.0cm of sis-text.west](rws) {weights};
    \node[fbox, draw=black, fit={(rws-v1) (rws-vK)}] (w1) {};
    \node[fbox, draw=blue, fit={(w1)}] (w2) {};
    \node[fbox, draw=teal, fit={(w2)}] (w3) {};
    \node[fbox, draw=orange, fit={(w3)}] (w4) {};
    \node[fbox, draw=purple, fit={(w4)}] (w5) {};
    \pic[right=1.5cm of rws-vK](nvi) {weights};
    \node[fbox, draw=black, fit={(nvi-v1)}] (w1) {};
    \node[fbox, draw=blue, fit={(w1) (nvi-v2)}] (w2) {};
    \node[fbox, draw=teal, fit={(w2) (nvi-v3)}] (w3) {};
    \node[fbox, draw=orange, fit={(w3) (nvi-vK-1)}] (wK-1) {};
    \node[fbox, draw=purple, fit={(wK-1) (nvi-vK)}] (wK) {};
    \draw[] (-2.3, -1.71) -- (13, -1.71);
    \node[align=left, below=1.0cm of sis-text.west, anchor=west](smc-text) {\small\textbf{\textsf{SMC}}};
    \pic[below=0.6cm of rws-v1](rws-rs) {weights};
    \node[fbox, draw=black, fit={(rws-rs-vK)}] (rws-rs-vK) {};
    \node[fbox, draw=blue, fit={(rws-rs-vK)}] (rws-rs-vK) {};
    \node[fbox, draw=teal, fit={(rws-rs-vK)}] (rws-rs-vK) {};
    \node[fbox, draw=orange, fit={(rws-rs-vK)}] (rws-rs-vK) {};
    \node[fbox, draw=purple, fit={(rws-rs-vK)}] (rws-rs-vK) {};
    \pic[below=0.6cm of nvi-v1](nvi-rs) {weights};
    \node[fbox, draw=black, fit={(nvi-rs-v1)}] (w1) {};
    \node[fbox, draw=blue, fit={(nvi-rs-v2)}] (w2) {};
    \node[fbox, draw=teal, fit={(nvi-rs-v3)}] (w3) {};
    \node[fbox, draw=orange, fit={(nvi-rs-vK-1)}] (wK-1) {};
    \node[fbox, draw=purple, fit={(nvi-rs-vK)}] (wK) {};
    \draw[line width=0.25mm] (-2.3, -2.7) -- (13, -2.7);
\end{tikzpicture}
    \caption{Weight contributions in the self-normalized gradient estimators for the forward KL-divergence for VI and NVI using SIS (no resampling) and SMC (resampling). 
    VI computes gradient estimates using the final weights (SIS), which simplify to the final incremental weight when resampling is performed (SMC). NVI computes gradient estimates based on the intermediate weights (SIS), which simplify to the intermediate incremental weights when resampling is performed (SMC).
    }
    \label{fig:nvi_weights}
\end{figure}

\section{Experiments}

\label{sec:experiments}
We evaluate NVI on three tasks,
(1) learning to sample form an unnormalized target density where intermediate densities are generated using annealing,
(2) learning heuristic factors to approximate the marginal likelihood of future observations in state-space models,
and finally 
(3) inferring distributions over classes from small numbers of examples in deep generative Bayesian mixtures.

\begin{figure}[t]
    \begin{subfigure}{\textwidth}
        \centering
        \includegraphics[width=\textwidth]{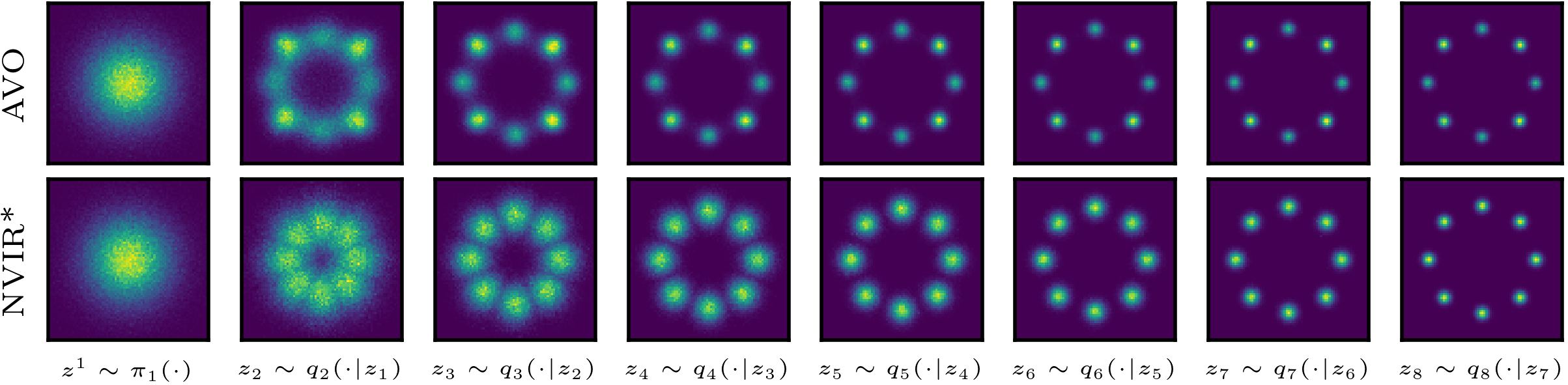}
        \label{fig:annealing_sample_paths_vdist}
    \end{subfigure}
    \begin{subfigure}{\textwidth}
        \includegraphics[width=\textwidth]{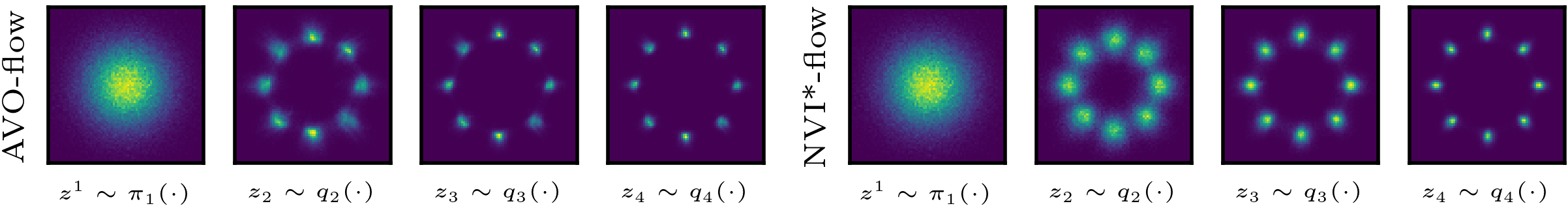}
        \label{fig:annealing_sample_paths_flow}
    \end{subfigure}
    \caption{
    \label{fig:annealing-gmm-avo-nvir}
    (\emph{Top}) Samples from forward kernels trained with AVO, and NVIR$^{*}$.
    (\emph{Bottom}) Samples from flow-based proposals trained with AVO-flow, and NVIR*-flow.
    }
\end{figure}

\subsection{Sampling from Multimodal Densities via Annealing}
\label{sec:exp-gmm-annealing}

A common strategy when sampling from densities with multiple isolated modes is to anneal from an initial density $\gamma_1$, which is typically a unimodal distribution that we can sample from easily, to the target density $\gamma_K$, which is multimodal \parencite{neal2001annealed}. Recent work on annealed variational objectives (AVOs) learns forward kernels $q_k$ and reverse kernels $r_{k-1}$ for an annealing sequence by optimizing a variational lower bound at each level of nesting \parencite{huang2018improving}, which is equivalent to NVI based on the reverse KL for a fixed sequence of densities $\gamma_k$, %
\begin{align}
    \max_{q_k, r_k}
    & \:\mathcal{L}^\textsc{avo}_k,
    &
    \mathcal{L}^\textsc{avo}_k
    &= 
    \Exp_{q_1, \dots, q_k}
    \big[ \log v_k \big],
    &
    \gamma_k(z) = q_1(z)^{1-\beta_k} \gamma_K(z)^{\beta_k}, 
    && k = 1, \ldots, K
    .
\end{align}
NVI allows us to improve upon AVO in two ways. First, we can perform importance resampling at every step to optimize an SMC sampler rather than an annealed importance sampler. 
Second, we can learn annealing paths $(\beta_1, \dots, \beta_K)$ which schedule intermediate densities $\gamma_k$ such that the individual KL divergences, and hence the expected log incremental importance weights, are minimized. 

We illustrate the effect of these two modifications in Figure~\ref{fig:annealing-gmm-avo-nvir}, in which we compare AVO to NVI with resampling and a learned path, which we refer to as NVIR$^*$. Both methods minimize the reverse KL at each step
\footnote{As noted in Section~\ref{sec:computing-gradients}, we found optimization of the forward KL to be unstable when learning $r_k$.}. 
For details on network architectures see Appendix~\ref{sec:apx_gmm_experiment}. The learned annealing path in NVIR$^*$ results in a smoother interpolation between the initial and final density. We also see that AVO does not assign equal mass to all 8 modes, whereas NVIR$^*$ yields a more even distribution.

\begin{figure}[t]
    \centering
    \includegraphics[width=\textwidth]{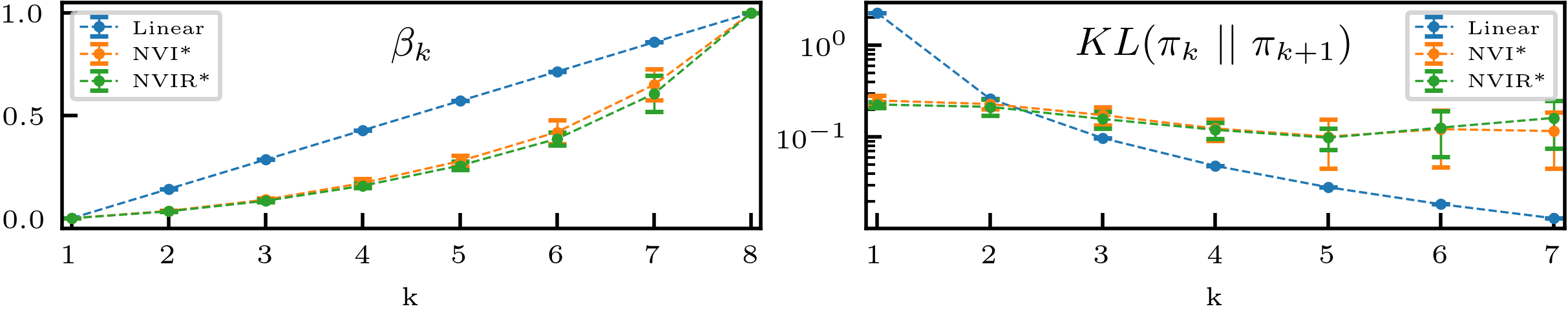}
    \caption{
    (\emph{Left}) Annealing paths learned by NVI$^{*}$ and NIVR* and the linearly spaced annealing geometric schedule (Linear) used by AVO, NVI, and NVIR. Results are averaged over 10 restarts; error bars indicate two standard deviations. 
    (\emph{Right}) The KL-divergences (computed by numeric integration) between consecutive intermediate distributions for different schedules.
    }
    \label{fig:annealing_schedules}
\end{figure}

\begin{figure}[t]
    \centering
    \includegraphics[width=\textwidth]{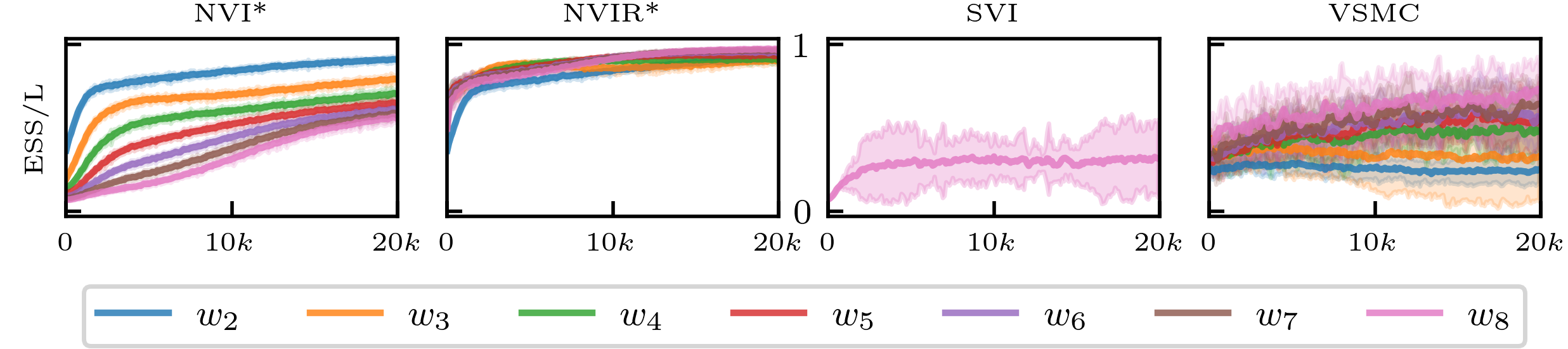}
    \caption{
       ESS relative to the number of samples $L=36$ during training for different methods using $7$ pairs of transition kernels (sequence length $K=8$) averaged across $10$ independent runs. Error bars indicate $\pm 2$ standard deviations; mean and standard deviation are computed based a rolling average with window size $100$.
    }
    \label{fig:training_ess}
\end{figure}

\begin{figure}[!t]
    \begin{minipage}[b]{0.60\textwidth}
        \scriptsize
        \begin{tabular}{lccccccccc}   
                \toprule
                &\multicolumn{4}{c}{$\mathbf{log\ \hat Z}$} ($\log Z \approx 2.08$)&
                &\multicolumn{4}{c}{\textbf{ESS}}\\
                \cmidrule{2-5}
                \cmidrule{7-10}
                Seq. length & K=2 & K=4 & K=6 & K=8&
                &K=2 & K=4 & K=6 & K=8
                \\
                \hline 
                SVI 
                & 1.86 & 1.89 & 1.92 & 1.72 &
                & 51 & 47 & 32 & 25
                \\
                SVI-flow 
                & 2.06 & - & - & - &
                & 55 & - & - & -
                \\
                \hline
                AVO
                & 1.86 & 1.96 & 2.01 & 2.05 &
                & 51 & 44 & 46 & 46
                \\
                NVI
                & 1.86 & 1.97 & 2.03 & 2.06 &
                & 51 & 45 & 45 & 41
                \\
                NVIR
                & 1.86 & 1.98 & 2.04 & 2.06 & 
                & 51 & \textbf{99} & \textbf{98} & \textbf{97}
                \\
                NVI$^{*}$
                & 1.86 & \textbf{2.06} & \textbf{2.07} & 2.07 & 
                & 51 & 51 & 54 & 54
                \\
                NVIR$^{*}$
                & 1.86 & \textbf{2.06} & \textbf{2.07} & \textbf{2.08} & 
                & 51 & 95 & 96 & \textbf{97}
                \\
                \hline
                AVO-flow
                & 2.05 & 2.08 & 2.07 & \textbf{2.08} &
                & 28 & 67 & 77 & \textbf{70}
                \\
                NVI*-flow
                & 2.05 & \textbf{2.08} & \textbf{2.08} & \textbf{2.08} &
                & 28 & \textbf{81} & \textbf{78} & \textbf{70}
                \\
                \bottomrule
            
            \end{tabular}
        \captionof{table}{
        Sample efficiency for NVI variants and baselines for $K-1$ annealing steps and $L$ samples per step for a fixed budget of $K \cdot L = 288$ samples. Metrics are computed for $100$ batches of $100$ samples per model across $10$ restarts.}
        \label{tab:quantitative_results_a}    
    \end{minipage}
    \hfill
    \begin{minipage}[b]{0.35\textwidth}
        \centering
        \includegraphics[height=1.4in]{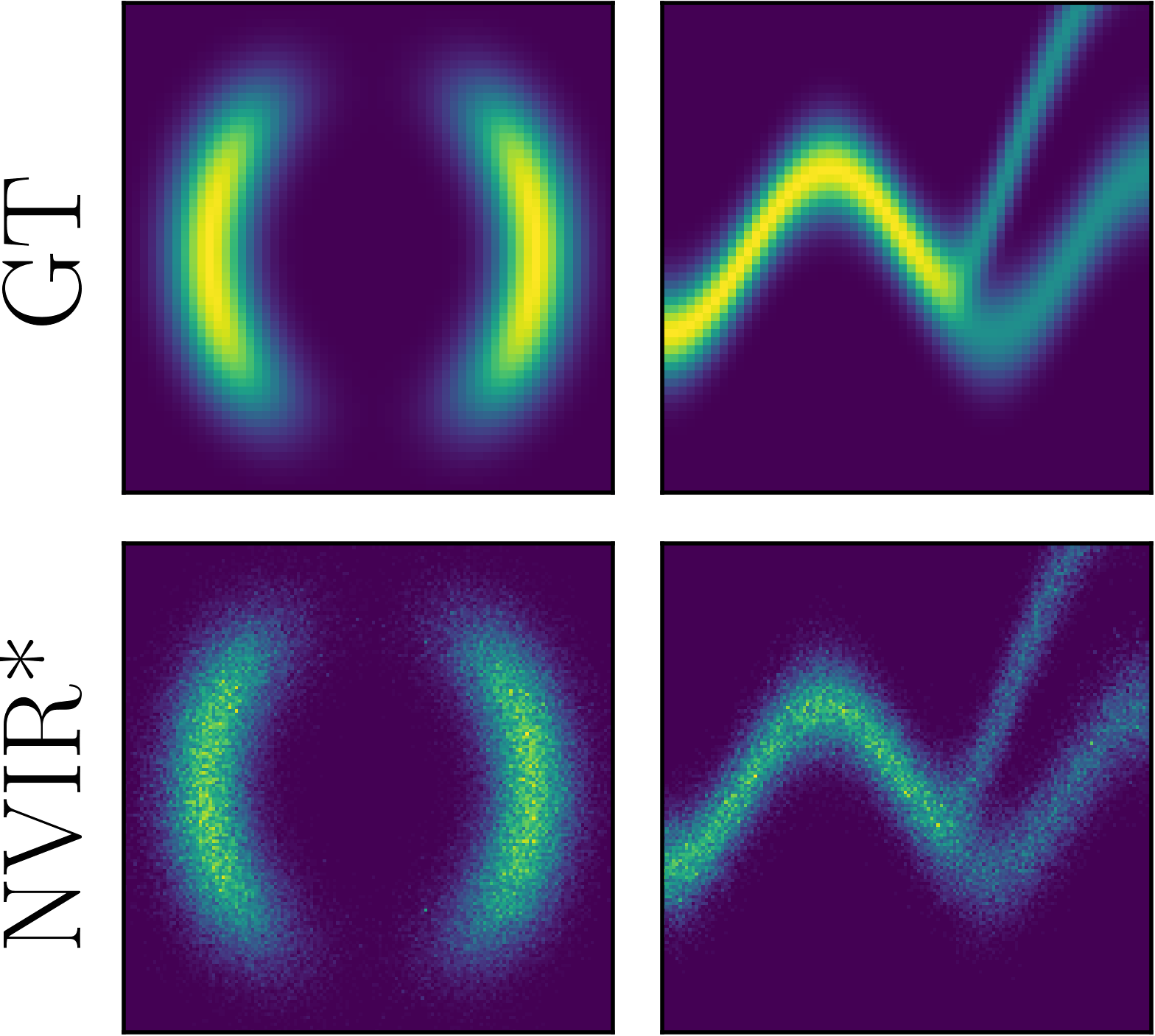}
        \captionof{figure}{Ground truth densities (GT) and samples from final target density for NVIR$^{*}$ with 2 intermediate densities.}
        \label{fig:flow_paper_densities}    
    \end{minipage}
\end{figure}

In Figure~\ref{fig:annealing_schedules} we compute the reverse KL between targets at each step in the sequence. For the standard linear annealing path, the KL decreases with $k$, suggesting that later intermediate densities are increasingly redundant. By constrast, in NVIR$^*$ we see that the KL is approximately constant across the path, which is what we would expect when minimizing $\mathcal{D}$ with respect to $\beta_k$. This is also the case in an ablation without resampling, which we refer to as NVI$^{*}$.

Figure~\ref{fig:training_ess} shows a rolling average of the ESS and its variance  during training. We compare NVI-based methods to SVI and a variational SMC sampler \parencite{le2018auto-encoding, maddison2017filtering, naesseth2017variational}.  
NVIR* has consistently higher ESS and significantly lower variance compared to baselines. These plots also provides insight into the role of resampling in training dynamics. In NVI$^{*}$, we observe a cascading convergence, which is absent in NVIR$^{*}$. We hypothesize that resampling reduces the reliance on high-quality proposals from step $k-1$ when estimating gradients at step $k$.

Annealed NVI has similar use cases as normalizing flows \parencite{rezende2015variational}. 
Inspired by concurrent work of \textcite{arbel2021annealed}, which explores a similar combination of SMC samplers and normalizing flows, we compare flow-based versions of NVI to planar normalizing flows, which maximize a standard lower bound (SVI-flow). 
We find that a normalizing flows can be effectively trained with NVI, in that samplers produce better estimates of the normalizing constant and higher ESS compared to SVI-flow. We also find that flow based models are able to produce high-quality samples with fewer intermediate densities (Figure~\ref{fig:annealing-gmm-avo-nvir}, bottom). Moreover, we see that combining a flow-based proposal with learned $\beta_k$ values (NVI$^{*}$-flow) results in a more accurate approximation of the target than in an ablation with a linear interpolation path (AVO-flow)\footnote{AVO-flow is itself a novel combination of AVO and flows, albeit an incremental one.}. 
resulting in a poor approximation to the target density (see Appendix~\ref{sec:apx_gmm_experiment}).

In Table \ref{tab:quantitative_results_a} we report sample quality in terms of the stochastic lower bound $\hat{\mathcal{L}}_K = \log \hat{Z}_K$ and the effective sample size $\text{ESS} = (\sum_{s} w^s_K)^2 / \sum_{s}(w^s_K)^2$. The first metric can be interpreted as a measure of the average sample quality, whereas the second metric quantifies sample diversity.
We compare NVI with and without resampling (NVIR$^{*}$ and its NVI$^*$) to ablations with a linear annealing path (NVIR and NVI), AVO, and a standard SVI baseline in which there are no intermediate densities. We additionally compare against AVO-flow and NVI$^{*}$-flow, which employ flows. We observe that NVIR$^{*}$ and NVI$^{*}$-flow outperform ablations and baselines in terms of $\log \hat{Z}$, and are competitive in terms of ESS. We show qualitative results for two additional target densities in Figure~\ref{fig:flow_paper_densities}.

\subsection{Learning Heuristic Factors for State-space Models}
\label{exp:state-space}

Sequential Monte Carlo methods are commonly used in state-space models to generate samples by proposing one variable at a time. To do so, they define a sequence of densities $\pi_k = \gamma_k/Z_k$ on the first $k$ time points in a model, which are also known as the filtering distributions,
\begin{align}
    \gamma_k (z_{1:k}, \eta)
    &= 
    p( x_{1:k}, z_{1:k}, \eta)
    =
    p(\eta) \: p(x_1, z_1 \,|\, \eta)
    \: \prod\nolimits_{l=2}^k \:
    p(x_l, z_l \,|\, z_{l-1}, \eta)
    ,
    &
    Z_k &= p(x_{1:k}).
\end{align}
Here $z_{1:k}$ and $x_{1:k}$ are sequences of hidden states and observations, and $\eta$ is a set of global variables of the model. These densities $\gamma_k$ differ from those in the annealing task in Section~\ref{sec:exp-gmm-annealing} in that the dimensionality of the support increases at each step, whereas all densities had the same support in the previous experiment. In this context, we can define a forward density $\hat{\gamma}_k$ that combines the preceding target $\gamma_{k-1}$ with a proposal $q_k$ for the time point $z_k$, and define a reverse density $\check{\gamma}_k = \gamma_{k}$ that is equal to the next intermediate density (which means that we omit $r_{k-1}$),
\begin{align}
    \check{\gamma}_k(z_{1:k}, \eta)
    &= 
    \gamma_k(z_{1:k}, \eta),
    &
    \hat{\gamma}_k(z_{1:k}, \eta)
    &=
    q_k(z_k \,|\, z_{1:k-1}) \:
    \gamma_{k-1} (z_{1:k-1}, \eta )
    .
\end{align}

A limitation of this construction is that the filtering distribution $\pi_{k-1}$ is not always a good proposal, since it does not incorporate knowledge of future observations. Ideally, we would like to define intermediate densities $\gamma_k(z_{1:k}, \eta) = p(x_{1:K}, z_{1:k}, \eta)$ that correspond to the smoothing distribution, but  this requires computation of the marginal likelihood of future observations $p(x_{k+1:K} \,|\, z_k, \eta)$, which is intractable. This is particularly problematic when sampling $\eta$ as part of the SMC construction. The first density $\pi_1(z_1, \eta) = p(z_1, \eta \,|\, x_1)$ will be similar to the prior, which will result in poor sampler efficiency, since the smoothing distribution $p(z_1, \eta \,|\, x_{1:K})$ will typically be much more concentrated.

To overcome this problem, we will use NVI to learn heuristic factors $\psi_\theta$ that approximate the marginal likelihood of future observations. We define a sequence of densities $(\gamma_0, \dots, \gamma_K)$,
\begin{align*}
    &
    \gamma_{0}(\eta) 
    = 
    p(\eta)
    \:
    \psi_\theta(x_{1:K} | \eta), 
    &&
    \gamma_{k} (z_{1:k}, \eta) 
    = 
    p(x_{1:k}, z_{1:k}, \eta)
    \:
    \psi_{\theta}(x_{k+1:K} \mid \eta),
    &&
    k= 1, 2, ..., K.
\end{align*}
Our goal is to learn parameters $\theta$ of the heuristic factor to ensure that intermediate densities approximate the smoothing distribution. This approach is similar to recently proposed work on twisted variational SMC \parencite{lawson2018twisted}, which maximized a stochastic lower bound.
\begin{figure}[!t]
    \centering
    \includegraphics[width=\textwidth]{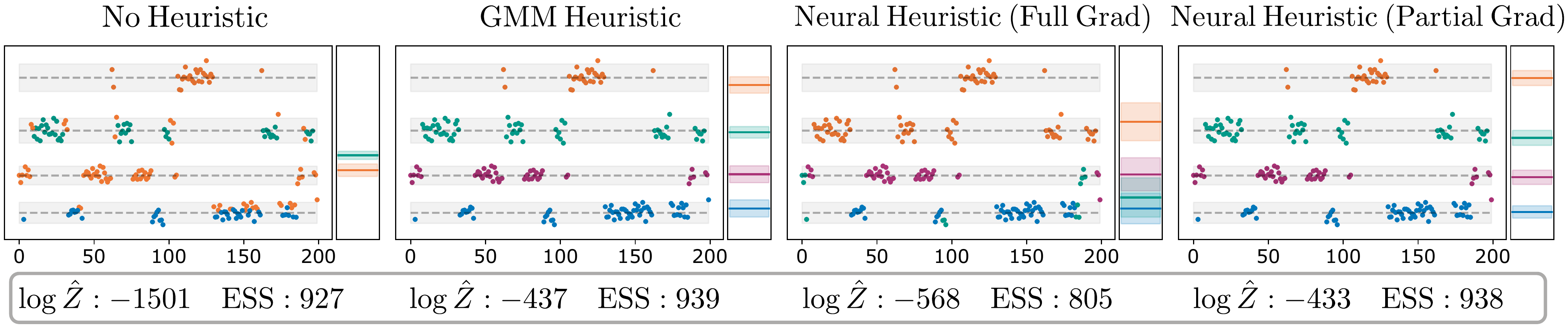}
    \caption{(\emph{Top}) qualitative results of 1 instance  with $K=200$ time steps (x-axis). Observations are color-coded based on the inferred assignments; Each colored band corresponds to the inferred cluster mean with one standard deviation, where the grey bands are the ground truth of the clusters. (\emph{Bottom}) We compute $\log \hat{Z}$ and ESS using 1000 samples and report average values on 2000 test instances.}
    \label{fig:hmm-samples}
\end{figure}

To evaluate the this approach, we will learn heuristic factors for a hidden Markov model (HMM). While HMMs are a well-understood model class, they are a good test case, in that they give rise to significant sample degeneracy in SMC and allow computation of tractable heuristics. We optimize an NVI objective based on the forward KL with respect to the heuristic factor $\psi_\theta$, an initial proposal $q_\phi(\eta \,|\, x_{1:K})$ and a forward kernel $q_\phi(z_k \,|\, x_k, z_{k-1}, \eta)$. We train on 10000 simulated HMM instances, each containing $K=200$ time steps and $M=4$ states. For network architectures see Appendix~\ref{sec:apx-exp-hmm}.

Figure~\ref{fig:hmm-samples} shows qualitative and quantitative results. We compare partial optimization with respect to $\hat{\gamma}_k$ to full optimization with respect to both $\hat{\gamma}_k$ and $\check{\gamma}_k$. We also compare NVI with neural heuristics to a baseline without heuristics, and a baseline that uses a Gaussian mixture model as a hand-coded heuristic. While full optimization yields poor results, partial optimization learns a neural heuristic whose performance is similar to the GMM heuristic, which is a strong baseline in this context.

\subsection{Meta Learning with Deep Generative models}
\label{subsec:bgmm-vae}

In this experiment, we evaluate NVI in the context of deep generative models with hierarchically-structured priors. Concretely, we consider the task of inferring class weights from a mini-batch of images in a fully unsupervised manner. For this purpose, we employ a  variational autoencoder with a prior in the form of a Bayesian Gaussian mixture model. We evaluate our model based on the quality of both the generative and inference model.

\textbf{Model Description.} We define a hierarchical deep generative model for batches of $N$ images of the form (see Appendix \ref{sec:apx-exp-bgmmvae} for a graphical model and architecture description)
\begin{equation*}
    \lambda \sim \text{Dir}(\cdot \,;\, \alpha)
    \quad 
    c_n \sim \text{Cat}(\cdot \,|\, \lambda)
    \quad
    z_n \sim \mathcal{N}(\cdot \,|\, \mu_{c_n},1/\tau_{c_n}) 
    \quad 
    x_n \sim p(\cdot \,|\, z_n; \theta)
    \quad
    \text{for } n = 1 \ldots N.
\end{equation*}
Here $\lambda$, $c_n$, $z_n$, and $x_n$ refer to the cluster probabilities, cluster assignments, latent codes, and observed images respectively. In the likelihood $p(x | z; \theta_x)$, we use a convolutional network to parameterize a continuous Bernoulli distribution~\parencite{NEURIPS2019_f82798ec}. We define proposals $q(z|x ; \phi_z)$, $q(c|z;\phi_c)$, and $q(\lambda|c;\phi_\lambda)$, which are also parameterized by neural networks. %
We refer to this model as Bayesian Gaussian mixture model VAE (BGMM-VAE). 

\textbf{Objective.} To construct an NVI objective, we define intermediate densities for $c$ and $z$. Unlike in previous experiments, we employ tractable densities in the form of a categorical  $\pi_c(c ; \theta_c)$ for cluster assignments and a 8-layer planar flow  $\pi_z(z; \theta_z)$ for the latent codes. We minimize the forward KL for the first two levels of nesting and the reverse KL at the final level of nesting
\begin{equation*}
    \KL
        {p(\lambda) \, p(c|\lambda) \vphantom{\big|}}
        {\vphantom{\big|} \pi_c(c) \, q(\lambda|c)}
    +
    \KL
        {\pi_c(c) \, p(z|c) \vphantom{\big|}}
        {\vphantom{\big|} \pi_z(z) \, q(c|z)}
    +
    \KL
        {\hat{p}(x) \, q(z|x) \vphantom{\big|}}
        {\vphantom{\big|} \pi_z(z) \, p(x|z)},
\end{equation*}
where $\hat{p}(x)$ is an empirical distribution over mini-batches of training data. We optimize the first two terms with respect to $\pi$ and $q$, and the third term with respect $q$ only. 

Since the intermediate densities are tractable in this model, no nested importance sampling is required to optimize this nested objective; we can compute gradient estimates base on a (single) sample from $p(\lambda) p(c|\lambda)$ in the first term, $\pi_c(c) q(\lambda|c)$ in the second, and $\hat{p}(x) q(z|x)$ in the final term. To learn the parameters $\{\mu, \tau, \theta_x\}$ of the generative model, we maximize a single-sample approximation of a lower bound $\mathcal{L} = \Exp_{q}\big[\log \big(p(x,z,c,\lambda)\,/\,q(z,c,\lambda|x)\big)\big].$

\newcolumntype{C}[1]{>{\centering\arraybackslash}m{#1}}
\begin{figure}[!t]
\centering
\begin{tabular}{C{0.05cm}C{6cm}C{6cm}}
    \rotatebox{90}{RWS} 
    &
    \includegraphics[width=0.45\textwidth]{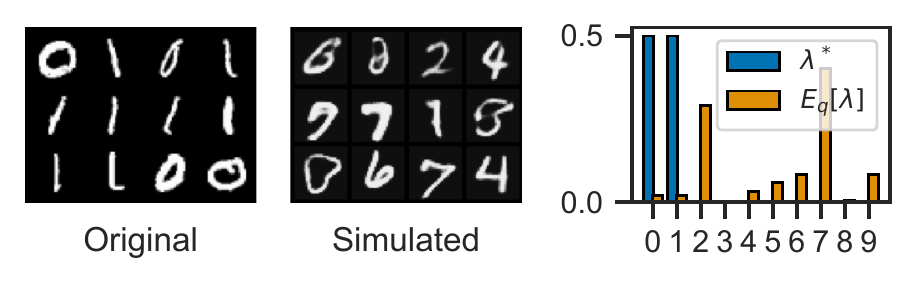}
    & 
    \includegraphics[width=0.45\textwidth]{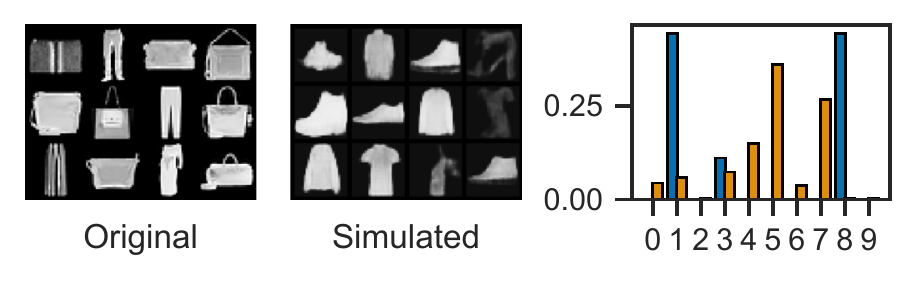}
    \\
    \rotatebox{90}{NVI} 
    & \includegraphics[width=0.45\textwidth]{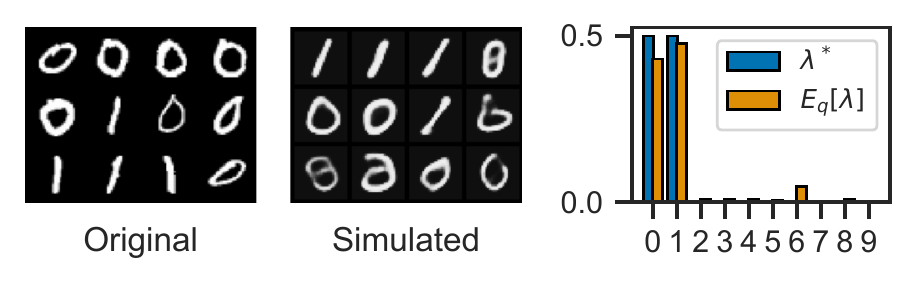}
    &
    \includegraphics[width=0.45\textwidth]{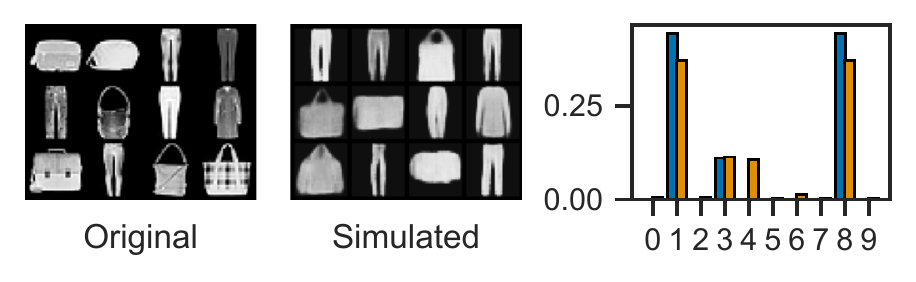} \\
\end{tabular}
    \caption{BGMM-VAE trained on MNIST \& FashionMNIST with the RWS objective (\emph{Top}) and the NVI objective (\emph{Bottom}). (\emph{Left}) Samples from a test mini-batch of size $N=300$. (\emph{Middle}) Samples from the generative model, generated from the $\lambda$ inferred from the test mini-batch.  (\emph{Right}) Comparison of ground truth $\lambda^*$ and the expected inferred value.}
    \label{fig:bgmm-vae-results}
\end{figure}

\textbf{Results.} We evaluate NVI for the BGMM-VAE using the following procedure. We generate mini-batches with a sampled $\lambda^*$ (for which we make use of class labels that are not provided to the model). We then compute the expectation of $\lambda$ under $q(\lambda,c,z|x)$ by sampling from the inference model, and compare this value against $\lambda^*$. Additionally, we generate a new mini-batch given the inferred $\lambda$ by running the generative model forward. We compare NVI against RWS, where we use 10 samples to estimate the gradients at each iteration. The results are shown in Figure~\ref{fig:bgmm-vae-results}. The cluster indices are rearranged based on the association of clusters to true classes. We observe that for RWS, even though the sample quality is reasonable, the posterior approximation is quite poor. When we train with NVI however, both the inferred $\lambda$ and the generated samples match the test mini-batch reasonably well.

\section{Conclusion}
\label{sec:conclusion}

We developed NVI, a framework that combines nested  importance  sampling  and  variational inference by optimizing a  variational objective at every level of nesting.
This formulation allows us to learn proposals and intermediate densities for a general class of samplers, which admit most commonly used importance sampling strategies as special cases. 
Our experiments demonstrate that samplers trained with NVI are able to outperform baselines when sampling from multimodal densities, Bayesian state-space models, and hierarchical deep generative models. 
Moreover, our experiments show that learning intermediate distributions results in better samplers. 

NVI is particularly useful in the context of deep probabilistic programming systems. Because NVI can be applied to learn proposals for a wide variety of importance samplers, it can be combined with methods for inference programming that allow users to tailor sampling strategies to a particular probabilistic program. Concretely, NVI can be applied to nested importance samplers that are defined using a grammar of composable \emph{inference combinators} \parencite{stites2021learning}, functions that implement primitive operations which preserve proper weighting, such as using samples from one program as a proposal to another program, sequential composition of programs, and importance resampling.

\section*{Acknowledgements}
This work was supported by the Intel Corporation, the 3M Corporation, NSF award 1835309, startup funds from Northeastern University, the Air Force Research Laboratory (AFRL), and DARPA.

\printbibliography
\end{refsection}

\newpage
\appendix
\begin{refsection}
\section{Related Work}
\label{sec:apx-related-work}

This work draws on three lines of research. The first are importance sampling and Sequential Monte Carlo methods, many of which can be described as properly weighted nested importance samplers. The second are approaches that combine importance sampling with stochastic variational inference, either by maximizing a stochastic lower bound, or by minimizing the forward KL divergence using self-normalized estimators. Finally there are a multitude of approaches that combine stochastic variational inference with some form of forward and reverse kernel, often in the form of an MCMC transition operator. We will discuss the most directly relevant approaches in these three lines of work. 

\paragraph{Importance Samplers, SMC, and Proper Weighting.}

There is a vast literature on importance sampling methods, a full review of which is beyond the scope of this paper. For introductory texts on SMC methods we refer to \textcite{doucet2009tutorial} and \textcite{doucet2001sequential}. Much of this literature has focused on state-space models, which define a distribution over a sequence of time-dependent variables. In these models, SMC is typically used to approximate the filtering distribution at each time, which is also known as particle filtering in this context. 

A limitation of SMC methods for filtering is that resampling introduces degeneracy; particles will typically coalesce to a common ancestor in $\mathcal{O}(S \log S)$ steps \parencite{jacob2013path}. This has given a rise to a literature on backwards simulation methods, which perform additional computation to reduce sample degeneracy (see \textcite{lindsten2012use} for a review). A second challenge is the estimation of global parameters of the likelihood and the transition distribution, which has given rise to a literature on particle Markov Chain Monte Carlo (PMCMC) methods \parencite{andrieu2010particle,lindsten2012ancestor,kantas2015particle}. Much of this literature orthogonal to the contributions in this paper. Backward simulation and PMCMC updates preserve proper weighting, and could therefore be used in NVI, but we do not consider such approaches here. 

There is also a large literature on applications of importance sampling and SMC in contexts other than state space models. SMC and its extensions have been applied to non-sequential graphical models \parencite{naesseth2014sequential}, and are widely used to perform inference in probabilistic programs \parencite{murray2013bayesian,wood2014new,vandemeent_aistats_2015,rainforth_icml_2016,murray2018automated}. As we mention in the main text, two classes of methods that are directly relevant to the experiments in this paper are annealed importance sampling \parencite{neal2001annealed} and SMC samplers \parencite{chopin2002sequential}, which can be used to target an annealing sequence by either applying an MCMC transition operator at each step, or by defining a density on an extended space in terms of a forward and reverse kernel in the same manner as we do in this paper. 

Much of this literature can be brought under a common denominator using the framework of proper weighting \parencite{naesseth2015nested, naesseth_elements_2019}, which formalizes the requirements on weights that are returned by an importance sampler relative to the unnormalized density that the sampler targets. This defines an abstraction boundary, which makes it possible reason about the validity of operations at the current level of nesting in terms of the marginal of the target density at the preceding level of nesting, without having to consider the full density on the extended space for sampling operations that precede the current level. In the context of NVI, the implication of this is that we could in principle different sampling strategies at each level of nesting. We have left such approaches to future work. 

\paragraph{Combining Importance Sampling and Stochastic Variational Inference.} 

In recent years we have seen a large number of approaches that combine stochastic variational inference with importance sampling. Much of the early work in this space was motivated by the desire to define a tighter variational lower bound. Work by \textcite{burda2016importance} proposed to train variational autoencoders by maximizing a stochastic lower bound $\hat{\mathcal{L}}=\log \hat{Z}$, where $\hat{Z}$ is defined by using the encoder distribution as a proposal in an importance sampler. This bound can be further tightened by using SMC rather than sequential importance sampling to compute $\hat{Z}$, which yields a lower-variance estimator $\hat{Z}$ \parencite{le2018auto-encoding,naesseth2018variational,maddison2017filtering}. More generally, tighter bounds have been proposed by using thermodynamic integration \parencite{masrani2019thermo} and by defining a bounds on the predictive marginal likelihood at each step \parencite{chen2021monte}.

Methods that make use of importance-weighted stochastic lower bounds have proven well-suited to maximum likelihood estimation, but suffer from a poor signal-to-noise ratio when performing variational inference. \textcite{rainforth2018tighter} show that the signal-to-noise ratio for the gradient with respect to the proposal parameters deteriorates as we increase the number of samples, which means that it is generally inadvisable to optimize a stochastic lower bound $\hat{\mathcal{L}}$ to learn these parameters. This realization has led to the development of doubly-reparameterized estimators \parencite{tucker2018doubly}, which have since been generalized to hierarchical models and score function terms w.r.t.~distribution other than the sampling distribution \parencite{bauer2021generalized}.

The realization that stochastic lower bounds are poorly suited to variational inference has also lead to a resurgence of interest in methods that derive from reweighted wake sleep \parencite{bornschein2015reweighted,le2019revisiting}, which minimize the forward KL divergence. A recent example along these lines is our own work on amortized Gibbs samplers \parencite{wu2019amortized}, which learns proposals that approximate Gibbs kernels in an SMC sampler. This method is a special case of NVI based on the forward KL divergence at each level of nesting, in which the same target density is used at each level of nesting.
Optimizing the a forward KL divergence to learn better variational proposals was also explored in recent work by \textcite{jerfel2021variational} using a variational boosting approach which iteratively constructs a Gaussian mixture model proposal. 

Concurrent work by \textcite{arbel2021annealed} focuses on combining AIS and SMC samplers with normalizing flows, where the reverse kernel can be chosen based on the inverse mapping defined by the flow.
Similar to our work, which mainly focuses on stochastic transitions, this work optimizes a sequence of KL divergences and makes use of importance resampling between steps. However this work focuses on flow-based models and does not consider the task of learning intermediate densities.

\paragraph{Combining MCMC with Stochastic Variational Inference.}

There are a large number of approaches that combine variational inference with MCMC and/or learned forward and reverse kernels, which leads to approaches that have similar use cases to the variational methods for SMC samplers that we consider in our experiments. Early work in this space by \textcite{salimans2015markov} computes a lower bound using Hamiltonian dynamics and defines  importance weights in terms of learned forward and reverse kernels. This work ommits the Metropolis-Hastings correction step typically used with HMC due to the inability of computing the density of the transition operator.
Later work by \textcite{wolf2016variational} incorporates an MH correction, hereby ensuring convergence to the posterior.

Work by \textcite{hoffman2017learning} also initialized HMC with samples from a variational proposal but does not learn forward and reverse transition kernels; this work simply optimizes a lower bound w.r.t.~the initial variational distribution and use the samples from HMC only to train the generative model.
\textcite{caterini2018hamiltonian} combine time-inhomogeneous Hamiltonian dynamics within variational inference using an SMC sampler construction where the reverse kernel can be chosen optimally by making use of the deterministic Hamiltonian dynamics. \textcite{wang2018meta} develop a meta-learning approach in which samples from a ground-truth generative model serve to train variational distributions that approximate Gibbs kernels for the generative model.

Optimizing a step-wise objectives to learn forward and reverse kernels has also previously been proposed by \textcite{huang2018improving}. However, this work differs from a SMC sampler trained with NVI in that samples are proposed from the marginal of the forward kernels as opposed to the last target density. Here, due to the intractibility of the marginal density, we can not compute and incremental importance weight to perform resampling.

\section{Notation}
\begin{tabular}{ll}
\toprule
    $\pi_k(z_k)$ 
    & 
    $k$-th target density on $\cZ_k$
    \\
    $\gamma_k(z_k) = Z_k \pi_k(z_k)$ 
    & 
    $k$-th unnormalized target 
    \\
    $\check \pi_k(z_k, z_{k-1}) =  \pi_k(z_k)r_{k-1}(z_{k-1} \mid \z_k, \check \f_k)$ 
    & 
    $k$-th extended target on $\cZ_{k-1} \times \cZ_k$
    \\
    $\check \gamma_k(z_k, z_{k-1}) = Z_k \check \pi_k(z_k, z_{k-1})$ 
    & 
    $k$-th extended unnormalized proposal 
    \\ 
    $\hat \pi_k(z_k, z_{k-1})= \pi_{k-1}(z_{k-1})\q_k(z_{k} \mid \z_{k-1}, \hat \f_k)$ 
    & 
    $k$-th extended proposal on $\cZ_{k-1} \times \cZ_k$
    \\
    $\hat \gamma_k(z_k, z_{k-1}) = Z_k \hat \pi_k(z_k, z_{k-1})$ 
    & 
    $k$-th extended unnormalized proposal 
    \\
    $v_k =
    \frac{
        \check \gamma(z_{k}, z_{k-1})
    }{
        \hat \gamma(z_{k}, z_{k-1})
    }
    = 
    \frac{
        \gamma_k(z_k)r_{k-1}(z_{k-1} \mid \z_k, \check \f_k)
    }{
        \gamma_{k-1}(z_{k-1})\q_k(z_{k} \mid \z_{k-1}, \hat
    \f_k)
    }$ 
    &
    $k$-th incremental weight
    \\
    $\tilde v_k 
    = 
    v_k \frac{Z_{k-1}}{Z_k} 
    = 
    \frac{
        \pi_k(z_k)r_{k-1}(z_{k-1} \mid \z_k, \check \f_k)
    }{
        \pi_{k-1}(z_{k-1})\q_k(z_{k} \mid \z_{k-1}, \hat
    \f_k)
    }$ 
    & 
    $k$-th normalized incremental weight
    \\
    $w_k = \prod_{k'=1}^k v_{k'}$ 
    & 
    $k$-th cumulative weight
    \\
\bottomrule
\end{tabular}

\section{Important Identities}
\label{sec:identities}
\paragraph{Thermodynamic Identity:}
\begin{align*}
    \dd{\theta}
    \log Z_\theta
    =
    \frac{1}{Z_\theta}
    \dd{\theta} 
    \int_{\mathcal{Z}_\theta}dz\
    \g(z; \theta)
    =
    \int_{\mathcal{Z}_\theta}dz\
    \frac{
        \g(z; \theta)
        }{
        Z_\theta
    }
    \dd{\theta} 
    \log \g(z; \theta)
    =
    \Exp_{z \sim \pi(\cdot; \theta)}
    \left[
        \pp[\log \g]{\theta}
    \right]
    .
\end{align*}

\paragraph{Log-derivative trick a.k.a.~reinforce trick:}
\begin{align}
    \label{eq:reinforce-property}
    \dd{\theta} \pi(z; \theta)
    = \pi(z) \frac{1}{\pi(z; \theta)} \dd{\theta} \pi(z; \theta)
    = \pi(z) \dd{\theta} \log\pi(z; \theta)
\end{align}
Consequently, it holds that
\begin{align*}
    \Exp_{z \sim \pi(\cdot; \theta)} 
    \left[
        \dd{\theta} \log \pi(z; \theta)
    \right]
    =
    \int_{\mathcal{Z}}dz\ 
    \pi(z; \theta) 
    \dd{\theta} \log \pi(z; \theta)
    =
    \int_{\mathcal{Z}}dz\ 
    \dd{\theta} \pi(z; \theta)
    =
    \dd{\theta} 
    \int_{\mathcal{Z}}dz\ 
    \pi(z; \theta)
    = 
    0
\end{align*}

\paragraph{Fisher's Identity:}
\begin{align*}
    \grad_\theta \log p_\theta(x) = \int dz\ p_\theta(z \mid x) \dd{\theta} \log p_\theta(x, z)
\end{align*}

\section{Gradient estimation}
\label{sec:apx-all-grad-estimate}
To compute the gradient of the nested variational objective (NVO) we need to compute the gradients of the individual
terms $\mathrm{D}_f \left(\check\pi_{k} \midd \hat\pi_{k}\right)$ w.r.t. parameters $\check
\phi_k, \hat \phi_k, \theta_{k},$ and $\theta_{k-1}$,
\begin{align*}
    \dd[\mathcal{D}]{\hat \phi_k} 
    &= 
    \dd[\mathrm{D}_f \left(\check\pi_{k} \midd \hat\pi_{k}\right)]{\hat \phi_k}
    ,
    &
    \dd[\mathcal{D}]{\theta_k} 
    &= 
    \dd[\mathrm{D}_f \left(\check\pi_{k} \midd \hat\pi_{k}\right)]{\theta_{k}}
    +
    \dd[\mathrm{D}_f \left(\check\pi_{k+1} \midd \hat\pi_{k+1}\right)]{\theta_{k}}
    ,
    \\
    \dd[\mathcal{D}]{\check \phi_k} 
    &= 
    \dd[\mathrm{D}_f \left(\check\pi_{k} \midd \hat\pi_{k}\right)]{\check \phi_k}
    ,
    &
    \dd[\mathcal{D}]{\theta_{k-1}} 
    &= 
    \dd[\mathrm{D}_f \left(\check\pi_{k} \midd \hat\pi_{k}\right)]{\theta_{k-1}}
    +
    \dd[\mathrm{D}_f \left(\check\pi_{k-1} \midd \hat\pi_{k-1}\right)]{\theta_{k-1}}
    .
\end{align*}
In the following we are deriving the relevant gradient terms for the general case, i.e. using an f-divergence, and state the gradient of the reverse KL-divergence, i.e. $f(w)= - \log w$, and forward KL-divergence, i.e. $f(w) = w \log w$, as special cases. 

\subsection{Gradients for general f-divergences}
\label{sec:apx-grad-f-divergence}

\paragraph{Gradient w.r.t. parameters $\hat \phi_k$ of the forward kernel.}
Reparameterizing the sample $z_k \equiv z_k(\e_k; \hat\phi_k)$ allows us, under mild conditions \footnote{These condition are given by the Leibniz integration rules}, to interchange the order of integration and differentiation and compute path-wise derivatives
\begin{align*}
    \small
    &\dd{\hat\phi_{k}} 
    \mathrm{D}_f\left(\check\pi_k \midd \hat\pi_k \right)
    \\
    =&
    \Exp_{z_{k-1} \sim \pi_{k-1}}
    \left[
        \Exp_{\e_k \sim p_k}
        \left[
            \dd{\hat \phi_k}
            f \left(
                v_k \frac{Z_{k-1}}{Z_{k}}
            \right)
        \right]
    \right]
    \\
    =&
    \Exp_{z_{k-1} \sim \pi_{k-1}}
    \left[
        \Exp_{\e_k \sim p_k}
        \left[
            \pp[f]{w} \biggr\rvert_{w = v_k \frac{Z_{k-1}}{Z_k}}
            \frac{Z_{k-1}}{Z_{k}}
            \pp[v_k]{z_k}
            \pp[z_k]{\hat \f_k}
            \biggr\rvert_{z_k = z_k(\e_k; \phi_k)}
            +
            \pp[f]{w} \biggr\rvert_{w = v_k \frac{Z_{k-1}}{Z_k}}
            \frac{Z_{k-1}}{Z_{k}}
            \pp[v_k]{\hat \f_k}
        \right]
    \right]
    \\
    =&
    \Exp_{z_{k-1} \sim \pi_{k-1}}
    \left[
        \Exp_{\e_k \sim p_k}
        \left[
            \pp[f]{w} \biggr\rvert_{w = v_k \frac{Z_{k-1}}{Z_k}}
            \frac{Z_{k-1}}{Z_{k}}
            \left(
                \pp[v_k]{z_k}
                \pp[z_k]{\hat \phi_k}
                \biggr\rvert_{z_k = z_k(\e_k; \phi_k)}
                -
                \pp[q_k]{\hat \phi_k}
            \right)
        \right]
    \right]
    \\
    =&
    \Exp_{z_{k-1} \sim \pi_{k-1}}
    \left[
        \Exp_{\e_k \sim p_k}
        \left[
            \pp[f]{w} \biggr\rvert_{w = v_k \frac{Z_{k-1}}{Z_k}}
            v_k
            \frac{Z_{k-1}}{Z_{k}}
            \left(
                \pp[\log v_k]{z_k}
                \pp[z_k]{\hat \phi_k}
                \biggr\rvert_{z_k = z_k(\e_k; \phi_k)}
                -
                \pp[\log q_k]{\hat \phi_k}
            \right)
        \right]
    \right]
    .
\end{align*}
Alternatively, we can compute a score function gradient which does not require the target density $\g_k$ to be differentiable w.r.t. the sample $z_k$ and
hence can also be computed for discrete variable models.
\begin{align*}
    \small
    &\dd{\hat\phi_{k}} 
    \mathrm{D}_f\left(\check\pi_k \midd \hat\pi_k \right)
    \\
    =&
    \Exp_{z_{k-1} \sim \pi_{k-1}}
    \left[
        \int_{\mathcal{Z}_k} dz_k\ 
            \dd{\hat{\f}_k}
            \left(
                q_k(z_k\mid z_{k-1}, \hat \f_k)
                f \left(
                    v_k \frac{Z_{k-1}}{Z_{k}}
                \right)
            \right)
    \right]
    \\
    =&
    \Exp_{z_{k-1} \sim \pi_{k-1}}
    \left[
        \Exp_{z_k \sim q_k(\cdot \mid z_{k-1}, \hat \f_k)}
        \left[
            f \left(
                v_k \frac{Z_{k-1}}{Z_{k}}
            \right)
            \pp[\log q_k]{\hat \f_k}
            +
            \pp[f]{w} \biggr\rvert_{w = v_k \frac{Z_{k-1}}{Z_k}}
            \frac{Z_{k-1}}{Z_k}
            \pp[v_k]{\hat \f_k}
        \right]
    \right]
    \\
    =&
    \Exp_{z_{k-1} \sim \pi_{k-1}}
    \left[
        \Exp_{z_k \sim q_k(\cdot \mid z_{k-1}, \hat \f_k)}
        \left[
            f \left(
                v_k \frac{Z_{k-1}}{Z_{k}}
            \right)
            \pp[\log q_k]{\hat \f_k}
            +
            \pp[f]{w} \biggr\rvert_{w = v_k \frac{Z_{k-1}}{Z_k}}
            v_k 
            \frac{Z_{k-1}}{Z_k}
            \pp[\log v_k]{\hat \f_k}
        \right]
    \right]
    \\
    =&
    \Exp_{z_{k-1} \sim \pi_{k-1}}
    \left[
        \Exp_{z_k \sim q_k(\cdot \mid z_{k-1}, \hat \f_k)}
        \left[
            \left(
                f \left(
                    v_k \frac{Z_{k-1}}{Z_{k}}
                \right)
                -
                \pp[f]{w} \biggr\rvert_{w = v_k \frac{Z_{k-1}}{Z_k}}
                v_k
                \frac{Z_{k-1}}{Z_k}
            \right)
            \pp[\log q_k]{\hat \f_k}
        \right]
    \right]
\end{align*}

\paragraph{Gradient w.r.t. parameters $\check \phi_k$ of the reverse kernel.}
Computing the gradient w.r.t. parameters of the reverse kernel is straightforward as the expectation does not depend on parameters $\check \phi_k$,
\begin{align*}
    &\dd{\check\phi_{k}} 
    \mathrm{D}_f\left(\check\pi_k \midd \hat\pi_k \right)
    \\
    =&
    \Exp_{z_{k-1}, z_k \sim \hat \pi_{k}}
    \left[
        \dd{\check \phi_k}
        f \left(
            v_k \frac{Z_{k-1}}{Z_{k}}
        \right)
    \right]
    \\
    =&
    \Exp_{z_{k-1}, z_k \sim \hat \pi_{k}}
    \left[
        \pp[f]{w} \biggr\rvert_{w = v_k \frac{Z_{k-1}}{Z_k}}
        \frac{Z_{k-1}}{Z_{k}}
        \pp[v_k]{\check \f_k}
    \right]
    \\
    =&
    \Exp_{z_{k-1}, z_k \sim \hat \pi_{k}}
    \left[
        \pp[f]{w} \biggr\rvert_{w = v_k \frac{Z_{k-1}}{Z_k}}
        v_k
        \frac{Z_{k-1}}{Z_{k}}
        \pp[\log v_k]{\check \f_k}
    \right]
    \\
    =&
    \Exp_{z_{k-1}, z_k \sim \hat \pi_{k}}
    \left[
        \pp[f]{w} \biggr\rvert_{w = v_k \frac{Z_{k-1}}{Z_k}}
        v_k \frac{Z_{k-1}}{Z_{k}}
        \pp[\log r_{k-1}]{\check \f_k}
    \right]
    .
\end{align*}

\paragraph{Gradient w.r.t. parameters $\theta_{k-1}$ of the \emph{current proposal}.}
The gradient w.r.t. $\theta_{k-1}$ requires to compute a score-function style gradient and the computation of the gradient of the log normalizing constant $\log Z_{k-1}$.
\begin{align*}
    \small
    &\dd{\theta_{k-1}} 
    \mathrm{D}_f\left(\check\pi_k \midd \hat\pi_k \right)
    \\
    =&
    \Exp_{z_{k-1} \sim \pi_{k-1}}
    \left[
        \pp[\log \pi_{k-1}]{\theta_{k-1}}
        \Exp_{z_k \sim q_k(\cdot \mid z_{k-1}, \hat \f_k)}
        \left[
            f 
            \left(
                v_k \frac{Z_{k-1}}{Z_{k}}
            \right)
        \right]
        +
        \Exp_{z_k \sim q_k(\cdot \mid z_{k-1}, \hat \f_k)}
        \left[
            \pp{\theta_{k-1}}
            f 
            \left(
                v_k \frac{Z_{k-1}}{Z_{k}}
            \right)
        \right]
    \right]
    \\
    =&
    \Exp_{z_{k-1}, z_k \sim \hat \pi_{k}}
    \left[
        f 
        \left(
            v_k \frac{Z_{k-1}}{Z_{k}}
        \right)
        \pp[\log \pi_{k-1}]{\theta_{k-1}}
        -
        \pp[f]{w} \biggr\rvert_{w = v_k \frac{Z_{k-1}}{Z_k}}
        v_k \frac{Z_{k-1}}{Z_{k}}
        \pp[\log \pi_{k-1}]{\theta_{k-1}}
    \right]
    \\
    =&
    \Exp_{z_{k-1}, z_k \sim \hat \pi_{k}}
    \left[
        \left(
            f 
            \left(
                v_k \frac{Z_{k-1}}{Z_{k}}
            \right)
            -
            \pp[f]{w} \biggr\rvert_{w = v_k \frac{Z_{k-1}}{Z_k}}
            v_k \frac{Z_{k-1}}{Z_k}
        \right)
        \pp[\log \pi_{k-1}]{\theta_{k-1}}
    \right]
    \\
    =&
    \Exp_{z_{k-1}, z_k \sim \hat \pi_{k}}
    \left[
        \left(
            f 
            \left(
                v_k \frac{Z_{k-1}}{Z_{k}}
            \right)
            -
            \pp[f]{w} \biggr\rvert_{w = v_k \frac{Z_{k-1}}{Z_k}}
            v_k \frac{Z_{k-1}}{Z_k}
        \right)
        \left(
            \pp[\log \g_{k-1}]{\theta_{k-1}}
            -
            \pp[\log Z_{k-1}]{\theta_{k-1}}
        \right)
    \right]
    \\
    =&
    \Exp_{z_{k-1}, z_k \sim \hat \pi_{k}}
    \left[
        \left(
            f 
            \left(
                v_k \frac{Z_{k-1}}{Z_{k}}
            \right)
            -
            \pp[f]{w} \biggr\rvert_{w = v_k \frac{Z_{k-1}}{Z_k}}
            v_k \frac{Z_{k-1}}{Z_k}
        \right)
        \pp[\log \g_{k-1}]{\theta_{k-1}}
    \right]
    \\
     &-   
    \Exp_{z_{k-1}, z_k \sim \hat \pi_{k}}
    \left[
        \left(
            f 
            \left(
                v_k \frac{Z_{k-1}}{Z_{k}}
            \right)
            -
            \pp[f]{w} \biggr\rvert_{w = v_k \frac{Z_{k-1}}{Z_k}}
            v_k \frac{Z_{k-1}}{Z_k}
        \right)
    \right]
    \Exp_{z_{k-1} \sim \pi_{k-1}}
    \left[\vphantom{ \biggr\rvert_{w = v_k \frac{Z_{k-1}}{Z_k}} }
        \pp[\log \g_{k-1}]{\theta_{k-1}}
    \right]
    \\
    =&
    \Cov_{\hat \pi_{k}}
    \left[
        f 
        \left(
            v_k \frac{Z_{k-1}}{Z_{k}}
        \right)
        -
        \pp[f]{w} \biggr\rvert_{w = v_k \frac{Z_{k-1}}{Z_k}}
        v_k \frac{Z_{k-1}}{Z_k}
        ,\
        \pp[\log \g_{k-1}]{\theta_{k-1}}
    \right]
    \\
    =&
    \Cov_{\hat \pi_{k}}
    \left[\vphantom{ \biggr\rvert_{w = v_k \frac{Z_{k-1}}{Z_k}} }
        f 
        \left(
            v_k \frac{Z_{k-1}}{Z_{k}}
        \right)
        ,\
        \pp[\log \g_{k-1}]{\theta_{k-1}}
    \right]
    -
    \Cov_{\hat \pi_{k}}
    \left[
        \pp[f]{w} \biggr\rvert_{w = v_k \frac{Z_{k-1}}{Z_k}}
        v_k \frac{Z_{k-1}}{Z_k}
        ,\
        \pp[\log \g_{k-1}]{\theta_{k-1}}
    \right]
\end{align*}

\paragraph{Gradient w.r.t. parameters $\theta_k$ of the \emph{current target}}
The gradient w.r.t. $\theta_k$ requires to compute the gradient of the log normalizing constant $\log Z_k$. 
\begin{align*}
    \small  
    &\dd{\theta_{k}} 
    \mathrm{D}_f\left(\check\pi_k \midd \hat\pi_k \right)
    \\
    =&
    \Exp_{z_{k-1}, z_k \sim \hat \pi_{k}}
    \left[
        \dd{\theta_k}
        f \left(
            v_k \frac{Z_{k-1}}{Z_{k}}
        \right)
    \right]
    \\
    =&
    \Exp_{z_{k-1}, z_k \sim \hat \pi_{k}}
    \left[
        \pp[f]{w} \biggr\rvert_{w = v_k \frac{Z_{k-1}}{Z_k}}
        \frac{Z_{k-1}}{Z_{k}}
        \pp[v_k]{\theta_k}
    \right]
    \\
    =&
    \Exp_{z_{k-1}, z_k \sim \hat \pi_{k}}
    \left[
        \pp[f]{w} \biggr\rvert_{w = v_k \frac{Z_{k-1}}{Z_k}}
        v_k \frac{Z_{k-1}}{Z_{k}}
        \pp[\log \pi_k]{\theta_k}
    \right]
    \\
    =&
    \Exp_{z_{k-1}, z_k \sim \hat \pi_{k}}
    \left[
        \pp[f]{w} \biggr\rvert_{w = v_k \frac{Z_{k-1}}{Z_k}}
            v_k \frac{Z_{k-1}}{Z_{k}}
        \left(
            \pp[\log \g_k]{\theta_k}
            -
            \pp[\log Z_k]{\theta_k}
        \right)
    \right]
    \\
    =&
    \Exp_{z_{k-1}, z_k \sim \hat \pi_{k}}
    \left[
        \pp[f]{w} \biggr\rvert_{w = v_k \frac{Z_{k-1}}{Z_k}}
        v_k \frac{Z_{k-1}}{Z_{k}}
        \pp[\log \g_k]{\theta_k}
    \right]
    - 
    \Exp_{z_{k-1}, z_k \sim \hat \pi_{k}}
    \left[
        \pp[f]{w} \biggr\rvert_{w = v_k \frac{Z_{k-1}}{Z_k}}
        v_k \frac{Z_{k-1}}{Z_{k}}
    \right]
    \Exp_{z_{k} \sim \pi_{k}}
    \left[
        \pp[\log \g_k]{\theta_k}
    \right]
\end{align*}

\subsection{Gradients for the reverse KL-divergence ($f(w) = -\log(w)$)}
\label{sec:apx-grad-reverse-kl}
Building on the deviations for the general case derived in Appendix~\ref{sec:apx-grad-f-divergence}, we derive the gradients for the reverse KL-divergence as special cases by substituting $f(w) = -\log(w)$.

\paragraph{Gradient w.r.t. parameters $\hat \phi_k$ of the forward kernel:}
\ \\The reparameterized gradient takes the form
\begin{align}
    \dd{\hat\phi_{k}} 
    \mathrm{D}_{- \log w}\left(\check\pi_k \midd \hat\pi_k \right)
    =&
    \Exp_{z_{k-1} \sim \pi_{k-1}}
    \left[
        \Exp_{\e_k \sim p_k}
        \left[
            -\pp[\log v_k]{z_k}
            \pp[z_k]{\hat\phi_k}
            - \pp[\log q_k]{\hat\phi_k}
        \right]
    \right]
    \\
    =&
    \Exp_{z_{k-1} \sim \pi_{k-1}}
    \left[
        \Exp_{\e_k \sim p_k}
        \left[
            -\pp[\log v_k]{z_k}
            \pp[z_k]{\hat\f_k}
        \right]
    \right],
\end{align}
whereas the score function gradient takes the form 
\begin{align}
    \dd{\hat\phi_{k}} 
    \mathrm{D}_{- \log w}\left(\check\pi_k \midd \hat\pi_k \right)
    &=
    \Exp_{z_{k-1} \sim \pi_{k-1}}
    \left[
        \Exp_{z_k \sim q_k(\cdot \mid z_{k-1}; \hat\phi_k)}
        \left[
            \left(
                1 
                - \log 
                \left(
                    v_k
                        \frac{Z_{k-1}
                    }{
                        Z_k
                    }
                \right)
            \right)
            \pp[\log q_k]{\hat \phi_k}
        \right]
    \right]
    \\
    &=
    \Exp_{z_{k-1}, z_k \sim \hat \pi_k}
    \left[
        - \log 
        v_k
        \pp[\log q_k]{\hat \f_k}
    \right]
    .
\end{align}
The final equalities hold due to the reinforce property 
\begin{align*}
    \Exp_{\e_k \sim p_k}
    \left[
        \pp[\log q_k]{\hat\f_k} \biggr\vert_{z_k={z_k(\e, \phi)}}
    \right]
    =
    \Exp_{z_k \sim q_k(\cdot \mid z_{k-1}, \hat\f_k)}
    \left[
        \pp[\log q_k]{\hat\f_k}
    \right]
    =
    0
    .
\end{align*}

\paragraph{Gradient w.r.t. parameters $\check \phi$ of the reverse kernel}
\begin{align}
    \dd{\check\phi_{k}} 
    \mathrm{D}_{- \log w}\left(\check\pi_k \midd \hat\pi_k \right)
    &=
    \Exp_{z_{k-1}, z_k \sim \hat \pi_k}
    \left[
        - \pp[\log r_k]{\check \phi_k}
    \right]
    .
\end{align}
\paragraph{Gradient w.r.t. parameters $\theta_k$ of the \emph{current target}} 
\begin{align}
    \label{eq:grad-exc-sf-app}
    \dd{\theta_{k}} 
    \mathrm{D}_{- \log w}\left(\check\pi_k \midd \hat\pi_k \right)
    &=
    \Exp_{z_{k-1}, z_k \sim \hat \pi_k}
    \left[
        -
        \pp[\log \gamma_k]{\theta_k}
    \right]
    +
    \Exp_{z_k \sim \pi_k}
    \left[
        \pp[\log \gamma_k]{\theta_k}
    \right]
    .
\end{align}

\paragraph{Gradient w.r.t. parameters $\theta_{k-1}$ of the \emph{current proposal}} 
\begin{align}
    \dd{\theta_{k-1}} 
    \mathrm{D}_{- \log w}\left(\check\pi_k \midd \hat\pi_k \right)
    =
    \Cov_{\hat \pi_{k}}
    \left[
        -
        \log 
        v_k
        ,
        \pp[\log \g_{k-1}]{\theta_{k-1}}
    \right]
\end{align}

\subsection{Gradients for the forward KL-divergence ($f(w) = w \log(w)$)}
\label{sec:apx-grad-forward-kl}
First notice that 
\begin{align}
    \mathrm{D}_{w \log w}(\check\pi_k \midd \hat\pi_k)
    &=
    \Exp_{z_{k-1}, z_k \sim \hat\pi_{k}}
    \left[
        w
        \log w
    \right]
    \\
    &=
    \Exp_{z_{k-1}, z_k \sim \check\pi_{k}}
    \left[
        \log w
    \right]
    \\
    &=
    \Exp_{z_{k-1}, z_k \sim \check\pi_{k}}
    \left[
        - \log w^{-1}
    \right]
    \\
    &=
    \mathrm{D}_{- \log w}(\hat\pi_k \midd \check\pi_k)
    .
\end{align}
Hence the gradients for the forward KL-divergence follow by symmetry from the gradient of the reverse KL-divergence by
identifying the components $r_k, \pi_k$ and corresponding parameters $\hat \phi_k, \theta_k$ with the components of the forward density $q_k, \pi_{k-1}$ and parameters $\check \phi_k, \theta_{k-1}$ respectively.

\paragraph{Gradient w.r.t. parameters $\hat \phi_k$ of the forward kernel:}
\begin{align*}
    \dd{\hat\phi_k}
    \mathrm{D}_{w \log w}\left(\check\pi_k \midd \hat\pi_k \right)
    &=
    \Exp_{z_{k-1}, z_k \sim \check\pi_{k}}
    \left[
        -
        \pp[\log q_k]{\hat\phi_k} 
    \right]
\end{align*}

\paragraph{Gradient w.r.t. parameters $\check \phi_k$ of the reverse kernel:}
Note that the sample $z_{k-1}$ is assumed to be non-reparameterized. Hence we only state the score-function gradient for the forward KL-divergence.
\begin{align}
    \dd{\check\phi_{k}} 
    \mathrm{D}_{w \log w}\left(\check\pi_k \midd \hat\pi_k \right)
    &=
    \Exp_{z_{k-1}, z_k \sim \check \pi_k}
    \left[
        \log 
        v_k
        \pp[\log r_k]{\check \phi_k}
    \right]
    .
\end{align}

\paragraph{Gradient w.r.t. parameters $\theta_{k}$ of the \emph{current target}} 
\begin{align}
    \dd{\theta_{k}} 
    \mathrm{D}_{w \log w}\left(\check\pi_k \midd \hat\pi_k \right)
    =
    \Cov_{\check \pi_{k}}
    \left[
        \log 
        v_k
        ,
        \pp[\log \g_{k}]{\theta_{k}}
    \right]
\end{align}

\paragraph{Gradient w.r.t. parameters $\theta_{k-1}$ of the \emph{current proposal}}
\begin{align}
    \dd{\theta_{k-1}} 
    \mathrm{D}_{w \log w}\left(\check\pi_k \midd \hat\pi_k \right)
    &=
    \Exp_{z_{k-1}, z_k \sim \check \pi_k}
    \left[
        -
        \pp[\log \gamma_{k-1}]{\theta_{k-1}}
    \right]
    +
    \Exp_{z_k \sim \pi_{k-1}}
    \left[
        \pp[\log \gamma_{k-1}]{\theta_{k-1}}
    \right]
    .
\end{align}
Notice that the expectations of the gradients for the forward KL-divergence are w.r.t. the extended target density $\check \pi$ as opposed to the extended proposal $\hat \pi$ as it is the case for the reverse KL-divergence.

\subsection{Estimation of expectations w.r.t. intermediate target densities.}
When estimating an expectation of some function $h$ w.r.t. an intermediate extended proposal $\hat \pi_k$,
we can rewrite the expectation w.r.t.~properly weighted samples from the previous level of nesting using Definition~\ref{def:proper_weighting} 
\begin{align}
    \Exp_{z_{k-1}, z_k \sim \hat \pi_k}
    \left[
        h(z_{k-1}, z_k)
    \right]
    =
    \Exp_{w_{k-1}, z_{k-1} \sim \Pi_{k-1}}
    \left[
        \frac{w_{k-1}}{cZ_{k-1}}
        \Exp_{z_k \sim q_k(\cdot \mid z_{k-1}, \hat \f_k)}
        \biggl[
            h(z_{k-1}, z_k)
        \biggr]
    \right]
    .
\end{align}
Here, $\Pi_k$ denotes the probability density over weighted samples $(z_k, w_k)$ of the nested importance sampler from the $k$-th level of nesting. In our experiments we have $c=1$.
Similarly, we can rewrite expectation w.r.t.~intermediate extended target density $\check \pi_k$
\begin{align}
    \Exp_{z_{k-1}, z_k \sim \check \pi_k}
    \left[
        h(z_{k-1}, z_k)
    \right]
    &=
    \Exp_{z_{k-1}, z_k \sim \hat \pi_k}
    \left[
        \frac{Z_{k-1}}{Z_k} 
        v_k\, 
        h(z_{k-1}, z_k)
    \right]
    \\
    &=
    \Exp_{w_{k-1}, z_{k-1} \sim \Pi_{k-1}}
    \left[
        \frac{w_{k-1}}{cZ_{k-1}}
        \Exp_{z_k \sim q_k(\cdot \mid z_{k-1}, \hat \f_k)}
        \biggl[
            \frac{Z_{k-1}}{Z_k} 
            v_k\, h(z_{k-1}, z_k)
        \biggr]
    \right]
    \\
    &=
    \Exp_{w_{k-1}, z_{k-1} \sim \Pi_{k-1}}
    \left[
        \frac{w_{k-1}v_{k}}{cZ_{k}}
        \Exp_{z_k \sim q_k(\cdot \mid z_{k-1}, \hat \f_k)}
        \biggl[
            h(z_{k-1}, z_k)
        \biggr]
    \right]
    .
\end{align}
The resulting expressions can be approximated using a self-normalized estimator as stated in Equation~\ref{eq:self_normalized_is_estimator}.
In NVI, we assume that samples $(z_k, w_k) \sim \Pi_k$ are non-reparameterized and hence do not \emph{carry back} gradient to the previous level of nesting. In practice, these samples might be generated using a reparameterized forward kernel and hence their gradient has to be detached.

\newpage
\section{Experiment Details}
\subsection{Experiment 1: Annealing}
\label{sec:apx_gmm_experiment}

We are targeting an unnormalized Gaussian mixture model (GMM)
\begin{align*}
    \gamma_K(z_K) 
        =\sum_{m=1}^M \mathcal{N}(z_K ;\ \m_m,\ \sigma^2 I_{2\times 2}),
    &&
    \mu_m 
        =\left(
            r \sin\left(\frac{2m\pi}{M}\right), 
            r \cos\left(\frac{2m\pi}{M}\right)
        \right)
        ,
\end{align*}
with $M=8$ equidistantly spaced modes with variance $\sigma^2=0.5$ along a circle with radius $r=10$. 

We define the initial proposal density $q_1$ to be a multivariate normal with mean $0$ and standard deviation $5$ and model the transition kernels to be conditional normal, 
\begin{align*}
    q_k(z_k \mid z_{k-1})
    &= 
    \mathcal{N}(z_k ;\ \mu_k(z_{k-1}), \sigma_k(z_{k-1})^2 \mathcal{I}_{2\times2}),
    \\
    r_{k-1}(z_{k-1} \mid z_k) 
    &= \mathcal{N}(z_{k-1} ;\ \mu_k(z_k), \sigma_k(z_k)^2 \mathcal{I}_{2\times2})
    .
\end{align*}
with mappings for the mean $\mu$ and standard deviation $\sigma$ as follows:
\begin{align}
    \mu(z) = W_\mu^T\left(h(z) + z\right) + b_\mu,
    &&
    \sigma(z) = \mathrm{softplus}(W_\sigma^T h(z) + b_\sigma)
    &&
    h(z) = W_h^T z + b_h
    .
\end{align}
In the experiments the hidden layer consists of $50$ neurons, i.e. $h(z) \in \mathbb{R}^{50}$.
For the flow-based models the kernels are specified by the flow and hence fully deterministic. In this case the incremental importance weight simplifies to 
\begin{align}
    v_k 
    =
    \frac{
        \gamma_{k+1}(z_{k+1})
    }{
        q_k(z_{k+1})
    }
    =
    \frac{
        \gamma_{k+1}(z_{k+1})
    }{
        \gamma_k(z_k) \log |J_{f_k^{-1}}(z_{k+1})|
    }
    =
    \frac{
        \gamma_{k+1}(f_k(z_{k}))
    }{
        \gamma_k(z_k) \log |J_{f_k}(z_{k})|^{-1}
    },
\end{align}
where the mapping $f_k$ is a planar flow consisting of $32$ layers.
All methods are trained for $20,000$ iteration using the Adam optimizer with a learning rate of $1\mathrm{e}^{-3}$. 

\subsection{Experiment 2: Learning Heuristic Factors for state-space models}
\label{sec:apx-exp-hmm}
We evaluate NVI for a hidden Markov model (HMM) where the likelihood is given by a mixture of Gaussians (GMM),
\begin{align*}
    \label{eq:hmm-generative-model}
    \tau_m, \mu_m 
    &\sim 
    p(\tau, \mu)
    =
    \text{NormalGamma}(\alpha_0, \beta_0, \mu_0, \nu_0),
    &
    m &= 1, 2, ..., M,
    \\
    z_1 
    &\sim 
    p(z_1)
    =
    \text{Categorical}(\pi),
    \\
    z_k \, | \, z_{k-1} = m 
    &\sim
    p(z_k \mid z_{k-1})
    =
    \text{Categorical}(A_m),
    \\
    x_k \, | \, z_k = m 
    &\sim 
    p(x_k \mid z_k, \tau, \mu)
    =
    \text{Normal}(\mu_m, \sigma_m),
    &
    k &= 1, 2, ..., K.
\end{align*}

where $M$ is the number of clusters and $K$ is the number of time steps in a HMM instance; $z_{1:K}$ and $x_{1:K}$ are the discrete hidden states and observations respectively, $\eta:=\{\tau_{m}, \mu_{m}\}_{m=1}^M$ is the set of global variables, where $\tau_m$, and $\mu_m$ are precision and mean of the m-th GMM cluster respectively.
We choose $M=4$, $K=200$, and the hyperparameters as follows
\begin{align}
    \alpha_0 &= 8.0, 
    &
    \beta &= 8.0,
    &
    \mu_0 &= 0.0,
    &
    \nu_0 &= 0.001,
    &
    \pi &= (0.25, 0.25, 0.25, 0.25).
\end{align}
\begin{align*}
    A = 
    \begin{bmatrix}
    \frac{9}{10} & \frac{1}{30} & \frac{1}{30} & \frac{1}{30} \\
    \frac{1}{30} & \frac{9}{10} & \frac{1}{30} & \frac{1}{30} \\ 
    \frac{1}{30} & \frac{1}{30} & \frac{9}{10} & \frac{1}{30} \\
    \frac{1}{30} & \frac{1}{30} & \frac{1}{30} & \frac{9}{10} \\
    \end{bmatrix}
\end{align*}

To perform inference, we learn the following proposals in form of
\begin{align}
    q_\f(\mu_{1:M}, \tau_{1:M} \mid x_{1:K}) 
    &=
    \prod_{m=1}^M
    \text{NormalGamma}
    (
        \tilde{\alpha}_m,
        \tilde{\beta}_m,
        \tilde{\mu}_m,
        \tilde{\nu}_m  
    ),
    \\
    q_\f(z_{1} \mid x_{1}, \mu_{1:M}, \tau_{1:M})
    &=
    \text{Categorical}
    (
    \tilde{\pi}_1
    ),
    \\
    q_\f(z_k \mid x_k, z_{k-1}, \tau_{1:M}, \mu_{1:M})
    &=
    \text{Categorical}
    (
    \tilde{\pi}_k
    ).
\end{align}
We use the tilde symbol $(\,\tilde{}\,)$ to denote the parameters of the variational distributions that are the outputs of the neural networks.
We also use NVI to learn a heuristic factor $\psi_\theta (x_{k:K} \mid \tau, \mu)$ in form of

\begin{align*}
    \psi_\theta (x_{k:K} \mid \tau, \mu) 
    &= 
    \prod\nolimits_{l=k}^K 
    \sum\nolimits_{m=1}^M \mathrm{Normal}(x_l; \tau_m, \mu_m) 
    \psi^{\mathrm{MLP}}_\theta (z_l=m \mid \eta, x_l)
    .
\end{align*}
In addition we consider a baseline that uses a Gaussian mixture model as hand-coded heuristic in form of

\begin{align*}
    \psi_{\text{GMM}}(x_{k:K} \mid \tau, \mu) 
    &= 
    \prod\nolimits_{l=k}^K
        \sum\nolimits_{m=1}^M 
            \mathrm{Normal} (x_l; \tau_m, \mu_m)
            p(z_l=m),
    \\
\end{align*}
We optimize the parameters $\{\f, \theta\}$ by minimizing forward KL divergences defines as follows 
\begin{align*}
    \mathcal{L}_0 (\theta, \f)
    &=
    \mathrm{KL}
    \left(
        \check\pi_{0, \theta} (\eta) \midd q_\f (\eta | x_{1:T})
    \right)
    \\
    &=
    \mathrm{KL}
    \left(
        \pi_{0, \theta} (\eta) \midd q_\f (\eta | x_{1:T})
    \right),
    \\
    \mathcal{L}_1 (\theta, \f)
    &=
    \mathrm{KL}
    \left(
        \check\pi_{1, \theta}(z_{1}, \eta) \midd \hat\pi_{1, \theta}(z_1, \eta)
    \right)
    \\
    &=
    \mathrm{KL}
    \left(
        \pi_{1, \theta}(z_{1}, \eta) \midd \pi_{0, \theta}(\eta) q_{\f}(z_1 | x_1, \eta)
    \right),
    \\
    \mathcal{L}_k (\theta, \f)
    &=
    \mathrm{KL}
    \left(
        \check\pi_{k, \theta}(z_{1:k}, \eta) \midd \hat\pi_{k, \theta}(z_{1:k}, \eta)
    \right)
    \\
    &=
    \mathrm{KL}
    \left(
        \pi_{k, \theta}(z_{1:k}, \eta) \midd \pi_{k-1, \theta}(z_{1:k-1}, \eta) q_{\f}(z_t | x_t, z_{k-1}, \eta)
    \right),
    &
    k &= 2, 3, ..., K.
\end{align*}
In NVI, we learn heuristic factors $\psi_\theta$ that approximate the marginal likelihood of future observations. We define a sequence of densities $(\gamma_0, \dots, \gamma_K)$,
\begin{align*}
    &
    \gamma_{0}(\eta) 
    = 
    p(\eta)
    \:
    \psi_\theta(x_{1:K} | \eta), 
    &&
    \gamma_{k} (z_{1:k}, \eta) 
    = 
    p(x_{1:k}, z_{1:k}, \eta)
    \:
    \psi_{\theta}(x_{k+1:K} \mid \eta),
    &&
    k= 1, 2, ..., K.
\end{align*}
In practice we found that partial optimization (i.e. only taking gradient w.r.t the right hand side of each KL) yields better performance compared to the full optimization of the objective. We will derive the gradient for each case.

\subsubsection{Partial optimization.} 
We consider only taking gradient w.r.t. the right hand side of each KL divergence.

When $k=0$, we have
\begin{align}
    - \nabla_{\f} \mathcal{L}_0 (\f)
    &=
    - \nabla_{\f}
    \mathrm{KL}
    \left(
        \pi_{0} (\eta) \midd q_\f (\eta \mid x_{1:T})
    \right)
    =
    \mathbb{E}_{\pi_0}
    \left[
    \nabla \log q_\f(\eta \mid x_{1:T})
    \right]
    .
\end{align}

When $k= 1: K$, we have
\begin{align}
    - \nabla_{\f, \theta}
    \mathcal{L}_k (\f, \theta)
    &=
    - \nabla_{\f, \theta}
    \mathrm{KL}
    \left(
        \pi_k(z_{1:k}, \eta) \midd \pi_{k-1, \theta}(z_{1:k-1}, \eta) q_\f(z_t \mid x_k, z_{k-1}, \eta)
    \right)
    \\
    &= 
    \mathbb{E}_{\pi_k}
    \left[
    \nabla \log \pi_{k-1, \theta}(z_{1:k-1}, \eta) + \nabla \log q_\f(z_t \mid x_k, z_{k-1}, \eta)
    \right]
    \\
    &=
    \mathbb{E}_{\pi_k}
    \left[
    \nabla_{} \log \psi_{\theta} (x_{k:K} \mid \eta) + \nabla \log q_\f(z_t \mid x_k, z_{k-1}, \eta)
    \right]
    \\
    &
    - \mathbb{E}_{\pi_{k-1}} [\nabla \log \psi_{\theta} (x_{k:K} \mid \eta)]
\end{align}

\paragraph{Full optimization.} Now we consider taking gradient w.r.t. the full objective.

When $k=0$,
\begin{align}
- \nabla_{\f, \theta} \mathcal{L}_0 (\f, \theta)
    &=
    - \nabla_{\f, \theta}
    \mathrm{KL}
    \left(
        \pi_{0, \theta} (\eta) \midd q_\f (\eta \mid x_{1:K})
    \right)
    \\
    &=
    \mathbb{E}_{\pi_0} 
    \left[
    -\nabla \log \pi_{0, \theta}(\eta \mid x_{1:K}) + \nabla \log q_\f(\eta \mid x_{1:K})
    \right]
    \\
    &+
    \mathbb{E}_{\pi_0}
    \left[
    - \nabla \log \pi_{0, \theta} (\eta \mid x_{1:T}) 
        \left(
        \log \frac{p(\eta) \psi_{\theta}(x_{1:K} \mid \eta)} {q_\f(\eta \mid x_{1:K})}
        - \log Z_0
        \right)
    \right]
    , \\
    &=
    \mathbb{E}_{\pi_0} 
    \bigg[
    \nabla \log q_\f(\eta \mid x_{1:K})
    \bigg]
    -
    \mathbb{E}_{\pi_0}
    \left[
    \log \frac{p(\eta) \psi_{\theta}(x_{1:K} \mid \eta)}{q_\f(\eta \mid x_{1:K})}  \nabla \log \psi_{\theta}(x_{1:K} \mid \eta)
    \right]
    \\
    &+
    \mathbb{E}_{\pi_0}
    \left[
    \log \frac{p(\eta) \psi_{\theta}(x_{1:K} \mid \eta)}{q_\f(\eta \mid x_{1:K})}
    \right]
    \mathbb{E}_{\pi_0}
    \bigg[
    \nabla \log \psi_{\theta}(x_{1:K} \mid \eta)
    \bigg]
\end{align}

If $k= 1: K$,
\begin{align}
&
- \nabla_{\f, \theta}
    \mathcal{L}_k (\f, \theta)
    \\
    &=
    - \nabla_{\f, \theta}
    \mathrm{KL}
    \left(
        \pi_{k, \theta}(z_{1:k}, \eta) \midd \pi_{k-1, \theta}(z_{1:k-1}, \eta) q_\f(z_k \mid x_k, z_{k-1}, \eta)
    \right)
    \\
    &= 
    \mathbb{E}_{\pi_k}
    \left[
    - \nabla \log \frac{\pi_{k, \theta}(z_{1:k}, \eta)}{\pi_{k-1, \theta}(z_{1:k-1}, \eta) \, q_\f (z_k \mid x_k, z_{k-1}, \eta)}
    \right]
    \\
    &+
    \mathbb{E}_{\pi_k}
    \left[
    \log \frac{\pi_{k, \theta}(z_{1:k}, \eta)}{\pi_{k-1, \theta}(z_{1:k-1}, \eta) \, q_\f (z_k \mid x_k, z_{k-1}, \eta)}
    \,
    \left(- \nabla \log \pi_{k, \theta} (z_{1:k}, \eta) \right)
    \right]
    \\
    &= 
    \mathbb{E}_{\pi_k}
    \left[
    \nabla \log \psi_{\theta} (x_{k:K} \mid \eta) + \nabla \log q_\f(z_k \mid x_k, z_{k-1}, \eta)
    \right]
    - \nabla \log Z_{k-1}
    \\
    &+
    \mathbb{E}_{\pi_k}
    \left[
    \log \frac{\psi_\theta(x_{k+1:K} \mid \eta)}{\psi_\theta(x_{k:K} \mid \eta) \, q_\f(z_k \mid x_k, z_{k-1}, \eta)} 
    \,
    \left(- \nabla \log \pi_k(z_{1:k}, \eta) \right)
    \right]
    ,\\
    &= 
    \mathbb{E}_{\pi_k}
    \left[
    \nabla \log \psi_\theta(x_{k:K} \mid \eta) + \nabla \log q_\f(z_k \mid x_k, z_{k-1}, \eta)
    \right]
    -
    \mathbb{E}_{\pi_{k-1}}
    \left[
    \nabla \log \psi_\theta(x_{k:K} \mid \eta)
    \right]
    \\
    &+
    \mathbb{E}_{\pi_k}
    \left[
    \log \frac{\psi_\theta(x_{k+1:K} \mid \eta)}{\psi_\theta(x_{k:K} \mid \eta) \, q_\f(z_k \mid x_k, z_{k-1}, \eta)} 
    \,
    \left(- \nabla \log \psi_\theta(x_{k+1:K} \mid \eta)  \right)
    \right]
    \\
    &+
    \mathbb{E}_{\pi_k}
    \bigg[
    \nabla \log \psi_\theta(x_{k+1:K} \mid \eta)
    \bigg]
    \,
    \mathbb{E}_{\pi_k}
    \left[
    \log \frac{\psi_\theta(x_{k+1:K} \mid \eta)}{\psi_\theta(x_{k:K} \mid \eta) \, q_\f(z_k \mid x_k, z_{k-1}, \eta)} 
    \right]
\end{align}

\subsubsection{Architectures of the Proposals and Heuristic Factor}
For the neural proposals, we employ the neural parameterizations based on the neural sufficient statistics~\parencite{wu2019amortized}. We will discuss each of them in the following.

\paragraph{Proposal for the global variables.}
\begin{align*}
q_\f(\mu_{1:M}, \tau_{1:M} \mid x_{1:K}) = \prod_{m=1}^M \text{NormalGamma}(\tilde{\alpha}_m, \tilde{\beta}_m, \tilde{\mu}_m, \tilde{\nu}_m)
\end{align*}
We firstly feed each $x_k$ into a MLP to predict pointwise features, also known as neural sufficient statistics~\parencite{wu2019amortized}
\begin{center}
    \begin{tabular}{l}
    \toprule
    Input $x_k\in\mathbb{R}^1$
    \\
    \midrule
    FC 128. Tanh. \\
    \hline
    FC 4. Softmax. $(t_{k, 1}, t_{k,2}, t_{k,3}, t_{k,4})$
    \\
    \bottomrule
    \end{tabular}
    \label{arch-hmm-global-nss}
\end{center}
Then we aggregate over all points and compute the intermediate-level features for each of the clusters,
\begin{align*}
    H_m &= \left(\sum_{k} t_{k, m}, \sum_{k} t_{k,m} \, x_k, \sum_{k} t_{k,m} \, x_k^2 \right),
    &
    m &= 1, 2, 3, 4.
\end{align*}
We feed these features into a MLP to predict the parameters of the variational distribution,
\begin{center}
    \begin{tabular}{l|l|l|l}
    \toprule
    \multicolumn{4}{l}{Input $H_m\in \mathbb{R}^3$}
    \\
    \midrule
    FC 128. Tanh. & FC 128. Tanh. & FC 128. Tanh. & FC 128. Tanh. \\
    \hline
    FC 1. Exp(). ($\tilde{\alpha}_m$) & FC 1. Exp(). ($\tilde{\beta}_m$) & FC 1. ($\tilde{\mu}_m$) & FC 1. Exp(). ($\tilde{\nu}_m$) 
    \\
    \bottomrule
    \end{tabular}
    \label{arch-hmm-global-params}
\end{center}
where Exp() means that we exponentiate the corresponding output values.

\paragraph{Proposal for the initial state.}
\begin{align*}
    q_\f(z_{1} \mid x_{1}, \tau_{1:M}, \mu_{1:M}) = \text{Categorical} (\tilde{\pi}_1)
\end{align*}
We concatenate each $x_k$ with each of the cluster parameters and then predict logits as the assignments, followed by a softmax normalization.
\begin{table}[!h]
    \centering
    \begin{tabular}{l}
    \toprule
        Input $x_k\in\mathbb{R}^1, \: \mu_m\in\mathbb{R}^1, \: \tau_m\in\mathbb{R}^1_{+}$
        \\
    \midrule
    Concatenate$[x_k\, \: \mu_m, \: \tau_m]$\\
    \hline
    FC 128. Tanh. FC 1. ($\tilde{\pi}_{k, m}$)\\
    \bottomrule
    \end{tabular}
    \label{arch-gmm-local-nss}
\end{table}
Then we normalize the logits as
\begin{align*}
    \tilde{\pi}_k
    =
    \text{Softmax}
    \Big(
    \tilde{\pi}_{k, 1}, \tilde{\pi}_{k, 2}, \tilde{\pi}_{k, 3}, \tilde{\pi}_{k, 4}
    \Big).
\end{align*}

\newpage
\paragraph{Proposal for the forward transitional state.}
\begin{align*}
   q_\f(z_k \mid x_k, z_{k-1}, \tau_{1:M}, \mu_{1:M}) = \text{Categorical} (\tilde{\pi}_k) 
\end{align*}
This is similar to the initial proposal, except that we concatenate the previous state as the input.
\begin{table}[!h]
    \centering
    \begin{tabular}{l}
    \toprule
        Input $x_k\in\mathbb{R}^1, \: z_{k-1}\in\mathbb{R}^1, \: \mu_m\in\mathbb{R}^1, \: \tau_m\in\mathbb{R}^1_{+}$
        \\
    \midrule
    Concatenate$[x_k\, \: z_{k-1}, \: \mu_m, \: \tau_m]$\\
    \hline
    FC 128. Tanh. FC 1. ($\tilde{\pi}_{k, m}$)\\
    \bottomrule
    \end{tabular}
\end{table}
We then normalize the logits using a softmax, 
\begin{align*}
    \tilde{\pi}_k
    =
    \text{Softmax}
    \Big(
    \tilde{\pi}_{k, 1}, \tilde{\pi}_{k, 2}, \tilde{\pi}_{k, 3}, \tilde{\pi}_{k, 4}
    \Big)
    .
\end{align*}

\paragraph{Heuristic factor $\psi_\theta (x_{k:K} \mid \eta)$.}\ \\
The neural heuristic factor takes as input the concatenation of each point and each of the cluster parameters, and output the logits
\begin{table}[!h]
    \centering
    \begin{tabular}{l}
    \toprule
        Input $x_k\in\mathbb{R}^1, \: \mu_m\in\mathbb{R}^1, \: \tau_m\in\mathbb{R}^1_{+}$
        \\
    \midrule
    Concatenate$[x_k\, \: \mu_m, \: \tau_m]$\\
    \hline
    FC 128. Tanh. FC 1. ($\psi_\theta (z = m \mid x_k, \tau_m, \mu_k)$)\\
    \bottomrule
    \end{tabular}
\end{table}

\newpage
\paragraph{More Qualitative Results.} We show inference results for more HMM instances.
\begin{figure}[!h]
    \centering
    \includegraphics[width=\textwidth]{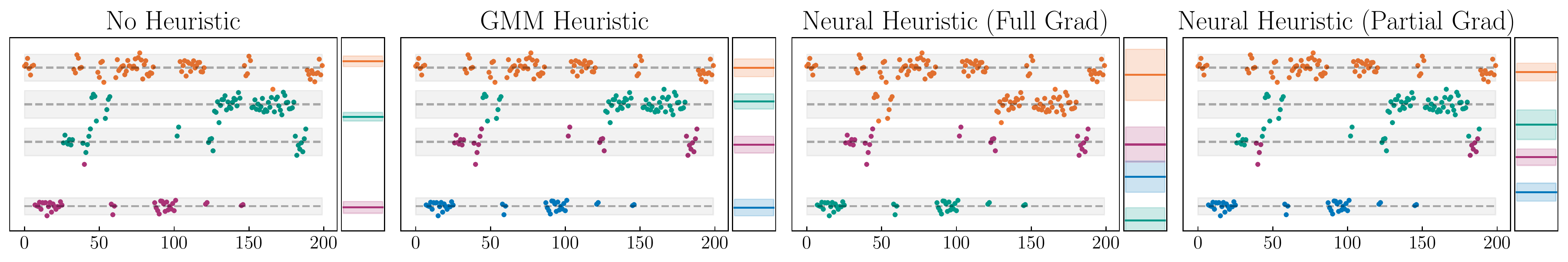}
    \includegraphics[width=\textwidth]{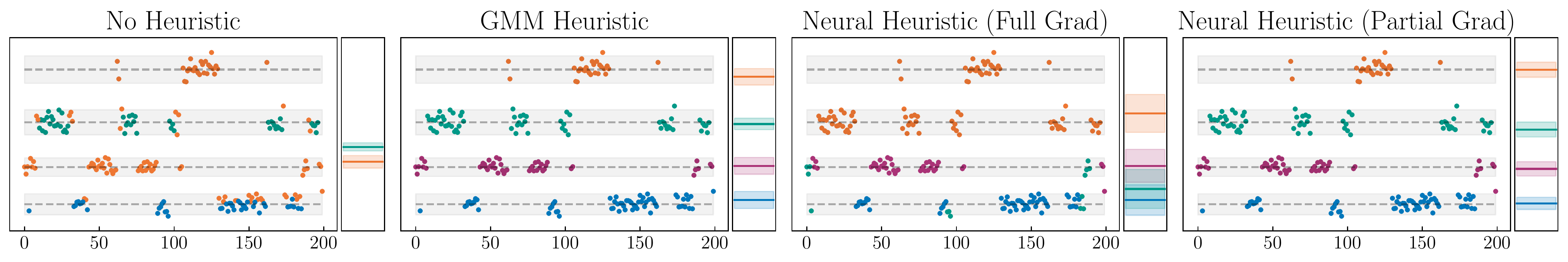}
    \includegraphics[width=\textwidth]{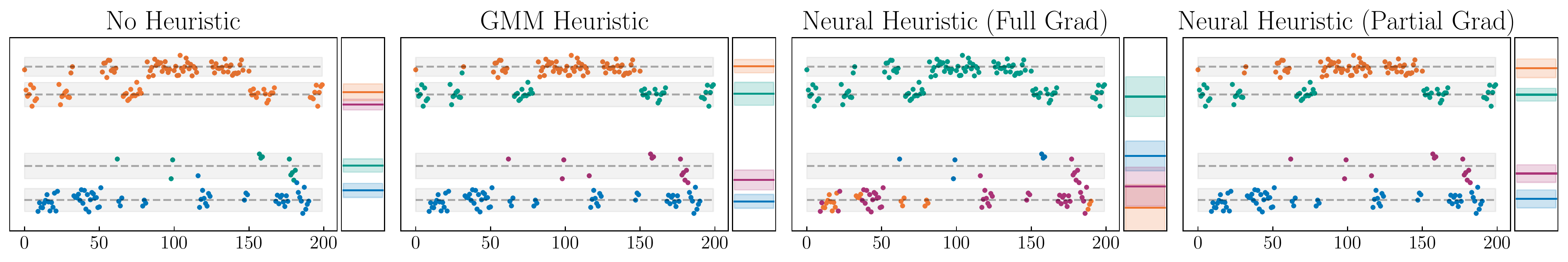}
    \includegraphics[width=\textwidth]{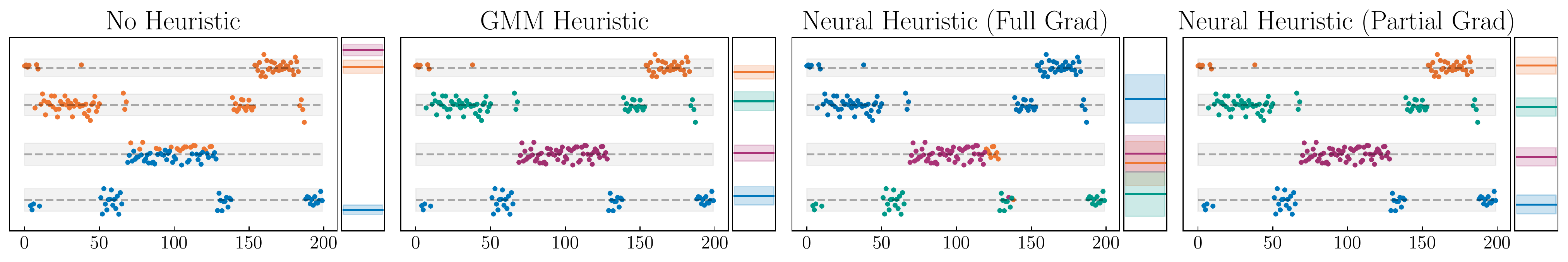}
    \includegraphics[width=\textwidth]{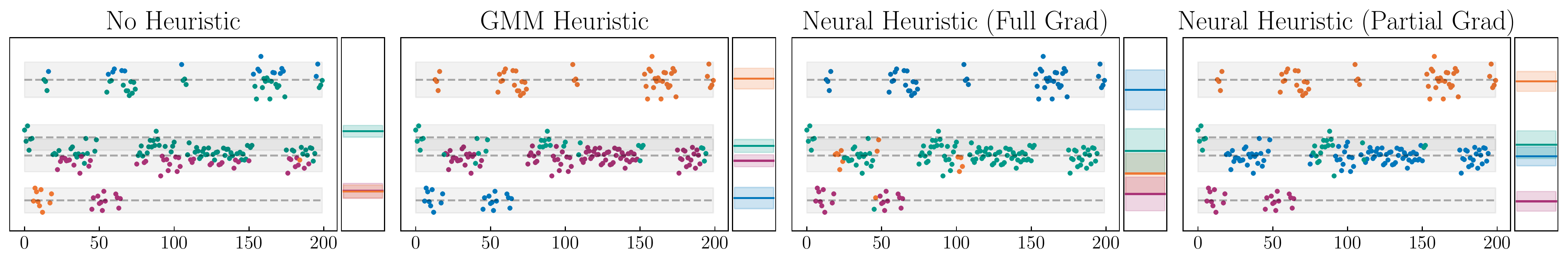}
    \includegraphics[width=\textwidth]{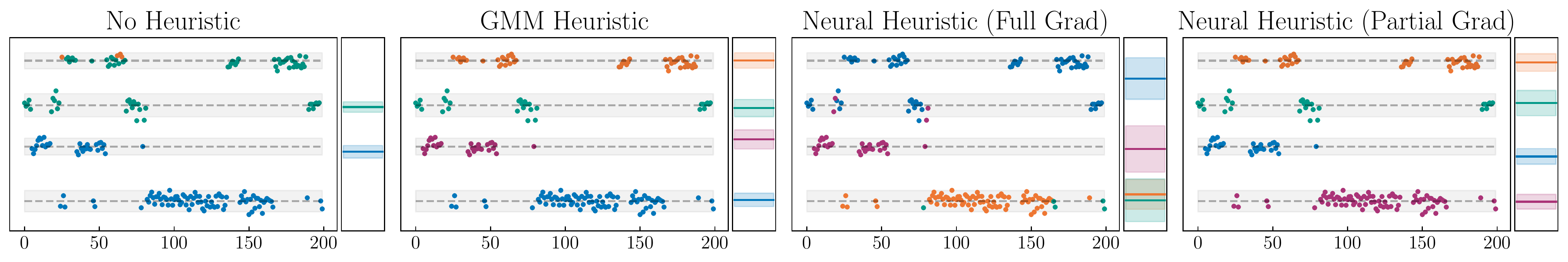}
    \includegraphics[width=\textwidth]{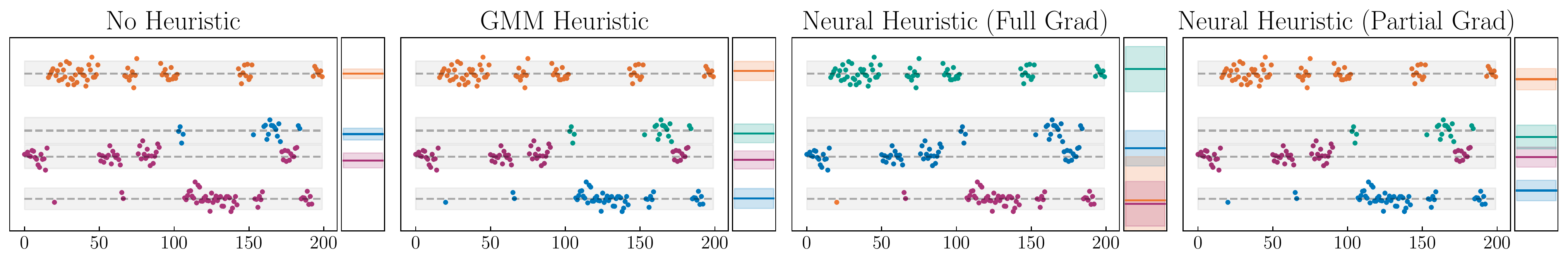}
    \includegraphics[width=\textwidth]{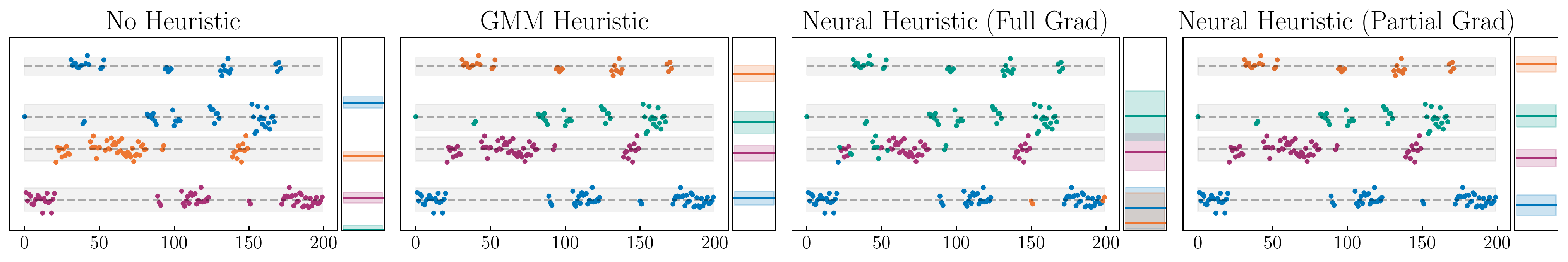}
    \includegraphics[width=\textwidth]{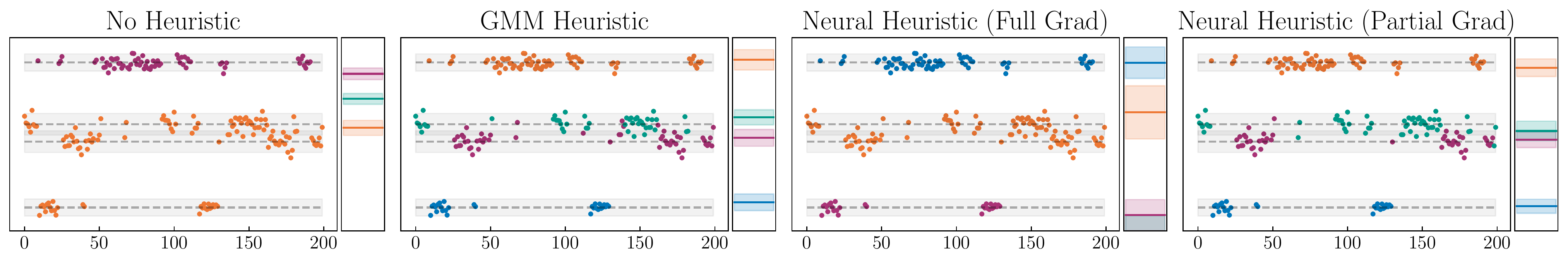}
    \label{fig:apx-hmm-more-results}
\end{figure}

\subsection{Experiment 3: BGMM-VAE}
\label{sec:apx-exp-bgmmvae}
We train on the datasets MNIST and FashionMNIST in which we sample mini-batches of size $N=10$ such that the classes are distributed based on a Dirichlet distribution with the same $\alpha$ as the generative model (We set $\alpha=0.5$ in our experiments). In NVI, we used a higher $N$ for the first two KLs (this is only applicable to NVI because the KLs are optimized locally).  For RWS, we used 10 samples per $x_n$. We trained all models for 50k iterations, and we used 20 mini-batches per iteration to estimate the overall objective. Fro the optimizers, we used Adam with learning rate 1e-3 for the $p(x|z,\theta_x)$ and $q(z|x,\phi_z)$ and 5e-2 for all models. 

\begin{figure}[!h]
\centering
\begin{tabular}{|c|c|}
\toprule
Generative Model & Inference Model \\
\midrule 
\begin{tikzpicture}[x=1.7cm,y=1.25cm]

\node[const] (alpha) {$\alpha$}; %

\node[latent, right=of alpha, xshift=-0.75cm] (pi) {$\lambda$}; %
\node[latent, right=of pi, xshift=-0.75cm] (c) {$c_n$}; %
\node[const, above=of c, yshift=-1.0cm] (mu) {$\mu$}; %
\node[const, right=of mu, xshift=-0.75cm] (tau) {$\tau$}; %

\node[latent, right=of c, xshift=-0.75cm] (z) {$z_n$}; %
\node[const, above=of z, yshift=-1.0cm, xshift=+0.5cm] (theta) {$\theta_x$}; %
\node[obs, right=of z,  xshift=-0.75cm]  (x) {$x_n$} ; %

\edge {alpha}{pi} ; %
\edge {pi}{c} ; %
\edge {c}{z} ; %
\edge {mu}{z} ; %
\edge {tau}{z} ; %
\edge {z}{x} ; %
\edge {theta}{x} ; %

\plate {plate2} { %
(c)(z)(x)
} {$N$}; %
\end{tikzpicture} 
&
\begin{tikzpicture}[x=1.7cm,y=1.25cm]

\node[latent, right=of alpha, xshift=-0.75cm] (pi) {$\lambda$}; %
\node[latent, right=of pi, xshift=-0.75cm] (c) {$c_n$}; %
\node[const, above=of x, yshift=-1.0cm] (phi_z) {$\phi_{z}$}; %
\node[const, above=of z, yshift=-1.0cm] (phi_c) {$\phi_{c}$}; %
\node[const, above=of c, yshift=-1.0cm] (phi_pi) {$\phi_{\pi}$}; %
\node[latent, right=of c, xshift=-0.75cm] (z) {$z_n$}; %
\node[obs, right=of z,  xshift=-0.75cm]  (x) {$x_n$} ; %

\edge {c}{pi} ; %
\edge {z}{c} ; %
\edge {x}{z} ; %

\edge {phi_z}{z} ; %
\edge {phi_c}{c} ; %
\edge {phi_pi}{pi} ; %

\plate {plate2} { %
(c)(z)(x)
} {$N$}; %
\end{tikzpicture} 
\\
\bottomrule
\end{tabular}
\caption{Overview of BGMM-VAE.}
\label{fig:bgmm-vae}
\end{figure}
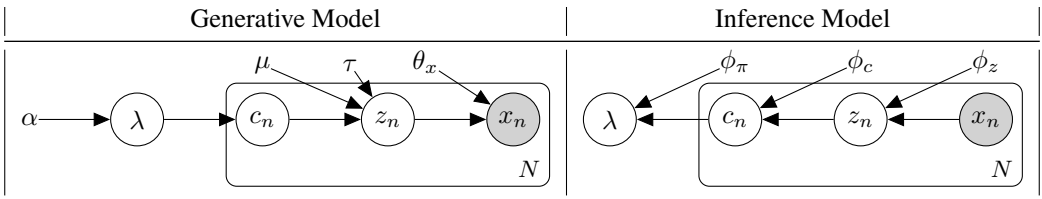

\begin{figure}[!h]
\centering
\begin{tabular}{l}
    \toprule
    \textbf{Encoder} $q(z|x;\phi_z)$
    \\
    \midrule
    \midrule
    Input $x\in\mathbb{R}^{1\times28\times 28}$
    \\
    \midrule
    Conv 32. $4 \times 4$, Stride 2, SiLU activation\\
    Conv 64. $4 \times 4$, Stride 2, SiLU activation\\
    Conv 64. $4 \times 4$, Stride 2, SiLU activation\\
    Conv 64. $4 \times 4$, Stride 2, SiLU activation\\
    FC $2 \times 10$
    \\
    \bottomrule
\end{tabular}
\vspace*{2ex}
\begin{tabular}{l}
    \toprule
    \textbf{Decoder} $p(x|z,\theta_x)$
    \\
    \midrule
    \midrule
    Input $z\in\mathbb{R}^{10}$
    \\
    \midrule
    FC 256. SiLU activation \\
    UpConv 64. $4 \times 4$, Stride 2, SiLU activation\\
    UpConv 64. $4 \times 4$, Stride 2, SiLU activation\\
    UpConv 64. $4 \times 4$, Stride 2, SiLU activation\\
    UpConv 32. $4 \times 4$, Stride 2, Sigmoid activation
    \\
    \bottomrule
\end{tabular}
\begin{tabular}{l}
    \toprule
    $q(c|z,\phi_c)$
    \\
    \midrule
    \midrule
    Input $z\in\mathbb{R}^{10}$
    \\
    \midrule
    FC 64. SiLU activation \\
    FC 64. SiLU activation \\
    FC 10 (\# clusters)\\
    Categorical(output) \\
    \bottomrule
\end{tabular}
\hspace*{6ex}
\begin{tabular}{l}
    \toprule
    $q(\pi|c,\phi_\pi)$
    \\
    \midrule
    \midrule
    Input $\{c_n\}_{n=1}^{N}\quad c_n\in\mathbb{N}^{+}$
    \\
    \midrule
    FC 64. SiLU activation \\
    Mean() (over $N$)\\
    FC 64. SiLU activation \\
    FC 10. SoftPlus activation (\# clusters)\\
    Dirichlet(output + $\sum_c$ c\_onehot) \\
    \bottomrule
\end{tabular}
\caption{BGMM-VAE architectures.}
\label{fig:bgmm-vae-archs}
\end{figure}

\printbibliography
\end{refsection}
\end{document}